\documentclass{article}
\usepackage{ragged2e}

\usepackage[preprint]{neurips_2025}

\usepackage[dvipsnames]{xcolor}
\usepackage[utf8]{inputenc} 
\usepackage[T1]{fontenc}    
\usepackage{hyperref}       
\usepackage{url}            
\usepackage{booktabs}       
\usepackage{amsfonts}       
\usepackage{nicefrac}       
\usepackage{microtype}      
\usepackage{microtype}
\usepackage{natbib}
\usepackage{graphicx}
\usepackage{subfigure}
\usepackage{booktabs} 
\usepackage[utf8]{inputenc} 
\usepackage[T1]{fontenc}    
\usepackage[textsize=tiny]{todonotes}
\usepackage{tcolorbox}
\usepackage{url}            
\usepackage{booktabs}       
\usepackage{amsfonts}       
\usepackage{nicefrac}       
\usepackage{microtype}      
\usepackage{adjustbox}
\usepackage{colortbl}
\usepackage{animate}
\usepackage{amssymb}
\usepackage{amsthm}
\usepackage{newtxtext}
\usepackage{bm}
\usepackage{bbm}
\usepackage{color}
\usepackage{booktabs}
\usepackage{multirow}
\usepackage{graphicx}
\usepackage{subcaption}
\usepackage{caption}
\usepackage{mathtools}
\usepackage{wrapfig}
\usepackage{bbding}

\usepackage{hyperref}

\usepackage{amsmath}
\usepackage{amssymb}
\usepackage{mathtools}
\usepackage{amsthm}

\usepackage[capitalize,noabbrev]{cleveref}

\usepackage{hyperref}
\usepackage{url}

\usepackage[utf8]{inputenc} 
\usepackage[T1]{fontenc}    
\usepackage{hyperref}       
\usepackage{url}            
\usepackage{booktabs}       
\usepackage{amsfonts}       
\usepackage{nicefrac}       
\usepackage{microtype}      
\usepackage{amssymb}
\usepackage{bbm}
\usepackage{amsmath}
\usepackage{graphicx}
\usepackage{multirow}
\usepackage{mathtools}
\usepackage{algorithm}
\usepackage{algorithmic}
\usepackage{enumerate}
\usepackage{wrapfig}
\usepackage{blindtext}
\usepackage{diagbox}
\usepackage{bigints}

\Crefname{section}{Sec.}{Secs.}
\Crefname{section}{Sec.}{Secs.}
\Crefname{appendix}{App.}{Apps.}
\Crefname{appendix}{App.}{Apps.}
\Crefname{figure}{Fig.}{Figs.}
\Crefname{figure}{Fig.}{Figs.}
\Crefname{table}{Tab.}{Tabs.}
\Crefname{table}{Tab.}{Tabs.}
\Crefname{algorithm}{Alg.}{Algs.}
\Crefname{algorithm}{Alg.}{Algs.}
\Crefname{equation}{Eq.}{Eqs.}
\Crefname{equation}{Eq.}{Eqs.}
\Crefname{lemma}{Lem.}{Lems.}
\Crefname{lemma}{Lem.}{Lems.}
\Crefname{theorem}{Thm.}{Thms.}
\Crefname{theorem}{Thm.}{Thms.}

\hypersetup{colorlinks, urlcolor=black, citecolor=Blue, linkcolor=ForestGreen}

\DeclareUnicodeCharacter{1D54F}{$\mathbb{X}$}
\DeclareUnicodeCharacter{1D569}{$\mathbbm{x}$}
\DeclareMathOperator*{\argmax}{arg\,max}

\DeclareMathOperator*{\sign}{sign}
\DeclareMathOperator*{\minimize}{minimize}
\DeclarePairedDelimiter\ceil{\lceil}{\rceil}

\newtheorem{theorem}{Theorem}
\newtheorem{lemma}{Lemma}
\newtheorem{corollary}{Corollary}
\newtheorem{assumption}{Assumption}

\newcommand{\boldit}[1]{\textbf{\textit{#1}}}

\newcommand{\insight}[2]{%
	\vspace{-0.18cm}%
	\begin{tcolorbox}[colback=gray!8!white,leftrule=2.5mm, boxrule=0.5mm, width=\textwidth+3mm, enlarge left by=-1mm, enlarge right by=-1mm, size=title]
		\textbf{#1}: #2
	\end{tcolorbox}
	\vspace{-0.16cm}%
}

\definecolor{cream}{RGB}{255,253,208}

\usepackage[textsize=tiny]{todonotes}

\title{Rewriting the Budget: A General Framework for Black-Box Attacks Under Cost Asymmetry}

\author{%
  Mahdi Salmani\textsuperscript{1} \quad
  Alireza Abdollahpoorrostam\textsuperscript{2} \quad
  Seyed-Mohsen Moosavi-Dezfooli\textsuperscript{3} \\
  \textsuperscript{1}University of Southern California \quad
  \textsuperscript{2}EPFL \quad
  \textsuperscript{3}Apple \\
  \texttt{salmanis@usc.edu} \quad
  \texttt{alireza.abdollahpoorrostam@epfl.ch} \quad
  \texttt{smoosavi@apple.com}
}

\begin{document}

\maketitle

\begin{abstract}
Traditional decision-based black-box adversarial attacks on image classifiers aim to generate adversarial examples by slightly modifying input images while keeping the number of queries low, where each query involves sending an input to the model and observing its output. Most existing methods assume that all queries have equal cost. However, in practice, queries may incur \textit{asymmetric costs}; for example, in content moderation systems, certain output classes may trigger additional review, enforcement, or penalties, making them more costly than others. While prior work has considered such asymmetric cost settings, effective algorithms for this scenario remains underdeveloped. In this paper, we propose a general framework for decision-based attacks under asymmetric query costs, which we refer to as \textit{asymmetric black-box attacks}.
We modify two core components of existing attacks: the \textit{search strategy} and the \textit{gradient estimation} process. Specifically, we propose \textit{Asymmetric Search~(AS)}, a more conservative variant of binary search that reduces reliance on high-cost queries, and \textit{Asymmetric Gradient Estimation~(AGREST)}, which shifts the sampling distribution to favor low-cost queries. We design efficient algorithms that minimize total attack cost by balancing different query types, in contrast to earlier methods such as \textit{stealthy attacks} that focus only on limiting expensive~(high-cost) queries. Our method can be integrated into a range of existing black-box attacks with minimal changes. We perform both theoretical analysis and empirical evaluation on standard image classification benchmarks. Across various cost regimes, our method consistently achieves lower total query cost and smaller perturbations than existing approaches, with improvements of up to 40\% in some settings. The code for \textit{Asymmetric Attacks} is available at \href{https://github.com/mahdisalmani/Asymmetric-Attacks}{github.com/mahdisalmani/Asymmetric-Attacks}.\looseness=-1 
\end{abstract}

\section{Introduction}
\label{sec:intro}

Decision-based adversarial attacks, first introduced by~\citep{brendel2017decision}, generate adversarial examples in black-box settings by systematically querying a classifier and observing only its output decisions for perturbed inputs. The original Boundary Attack~\citep{brendel2017decision} initially required over 100,000 queries to reliably identify minimal adversarial perturbations for large-scale datasets such as ImageNet~\citep{deng2009imagenet}. Subsequent works~\citep{chen2020hopskipjumpattack,chen2020raysraysearchingmethod,cheng2018query,cheng2019sign,qfool,rahmati2020geoda} significantly enhanced the efficiency of decision-based attacks by reducing the number of queries needed, achieving improvements of one to three orders of magnitude. These advancements have led to more practical and efficient frameworks for adversarial testing in limited-query settings.\looseness=-1

While prior work (discussed in detail in \Cref{app:rel_work}) has primarily assumed that all queries have equal cost and focused on minimizing the total number of queries, in many practical scenarios, queries can incur asymmetric costs depending on their nature. For instance, Not Safe for Work (NSFW) image detection models have become increasingly important, with major platforms such as Facebook~\citep{facebookDoesFacebook} and $\mathbb{X}$ (formerly Twitter)\citep{twitterSensitiveMedia} deploying automated mechanisms for identifying sensitive content, alongside commercial APIs developed by Google\citep{googleDetectExplicit}, Amazon~\citep{amazonModeratingContent}, and Microsoft~\citep{microsoftDetectAdult}. In these settings, submitting explicit or borderline explicit queries could trigger more severe consequences, such as account suspension or content flagging, compared to benign queries. As a result, minimizing only the total number of queries is insufficient; effective attack strategies must account for the asymmetric costs associated with different types of queries.\looseness=-1

\begin{wrapfigure}{r}{0.50\textwidth}
    \centering
    \vspace{-0.5cm}
    \includegraphics[width=\linewidth]{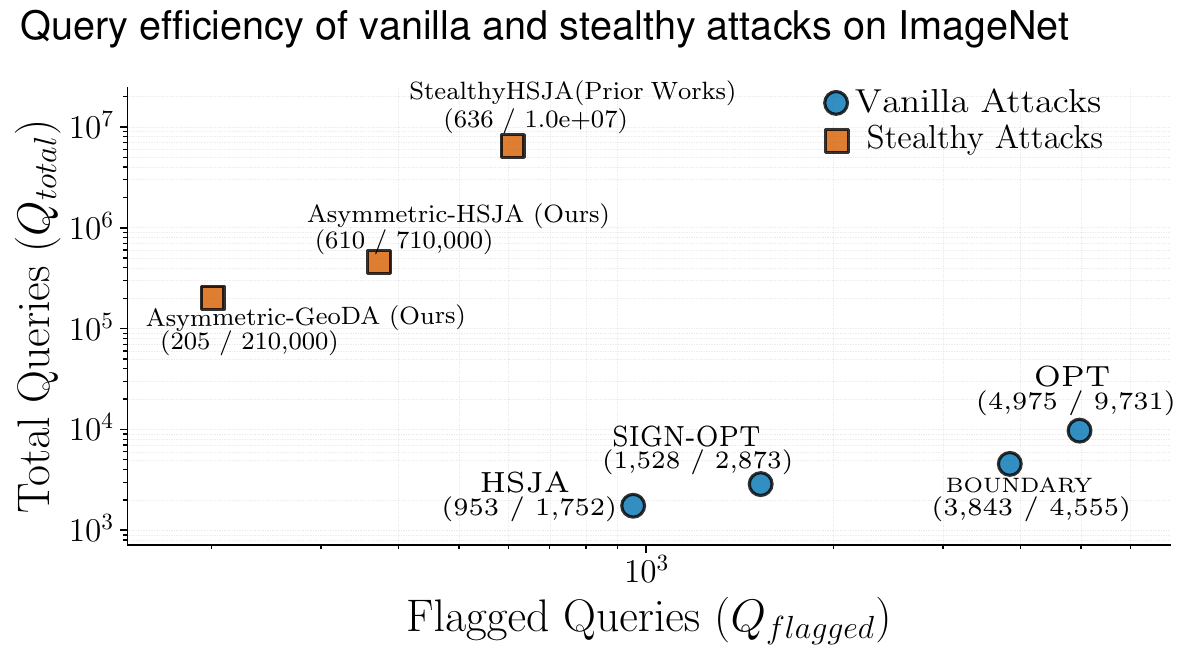}
    \vspace{-0.65cm}
    \caption{Each point represents the median number of queries required by an attack method to reach a median $\ell_2$ norm of 10. The x-axis shows the number of flagged queries ($Q_{\text{flagged}}$) and the y-axis reports the total number of queries ($Q_{\text{total}}$). 
    It demonstrates the superiority of our method in achieving a more favorable trade-off between flagged and total number of queries in stealthy attack settings.} 
    \vspace{-0.5cm}
    \label{fig:comparison_stealthy}
\end{wrapfigure}
Debenedetti et al.~\citep{debenedetti2024evading} introduced \textbf{\textit{stealthy attack}} techniques to better handle asymmetric query costs. They empirically demonstrated that the standard binary search procedure for boundary point detection, mostly for projecting an adversarial point onto the decision boundary or for OPT-style gradient estimation\citep{cheng2018query}, leads to a large number of high-cost queries. In particular, it can be observed from Fig.\ref{fig:comparison_stealthy} that at least 50\% of the queries made during these attacks are high-cost. To address this, they replaced the binary search with a search strategy inspired by the classic egg dropping problem~\citep{brilliantDroppingBrilliant}, which is similar to a line search algorithm. However, they did not provide a \textit{stealthy} variant of the Monte Carlo gradient estimation used in HSJA~\citep{chen2020hopskipjumpattack}, GeoDA~\citep{rahmati2020geoda}, and qFool~\citep{qfool}. Instead, they substituted it with an OPT-style gradient estimation procedure~\citep{cheng2018query}.\looseness=-1

Although stealthy attacks move toward addressing asymmetric query costs, they are not designed to handle arbitrary cost ratios. They implicitly assume that benign queries have zero cost, which may not reflect realistic settings where even benign queries contribute to the overall cost. In addition, since stealthy attacks could not adapt the Monte Carlo gradient approximation used in HSJA~\citep{chen2020hopskipjumpattack}, they instead rely on a suboptimal and inefficient OPT-style gradient estimation~\citep{cheng2018query}, which is already significantly outperformed by the HSJA gradient approximation under \textbf{\textit{symmetric}} cost settings. These limitations motivate us to answer the following question:\looseness=-1
\insight{Q}{How can we develop an \textbf{\textit{efficient}} framework to adapt attacks for any \textbf{\textit{arbitary}} cost ratio \textbf{\textit{without discarding}} any of their core components, including gradient estimation and binary search?\looseness=-1}

In this work, we propose a general framework for decision-based attacks that operates under arbitrary query cost asymmetries. Instead of minimizing only the number of high-cost queries, we adapt the core components of black-box attacks, namely search along adversarial paths and gradient estimation, to explicitly minimize the \textit{total query cost}. Our framework handles any cost ratio between high-cost and low-cost queries and completely outperforms stealthy attacks by optimizing the attack structure without sacrificing efficiency. Unlike stealthy attacks~\citep{debenedetti2024evading}, which modify the core mechanics of existing attacks and rely on inefficient gradient approximations, we retain the more efficient Monte Carlo gradient estimation technique used in HSJA~\citep{chen2020hopskipjumpattack}, GeoDA~\citep{rahmati2020geoda}, and qFool~\citep{qfool} with only slight adaptations to account for asymmetric costs.\looseness=-1

We first adapt the binary search procedure to account for arbitrary query costs. Instead of splitting the search interval into two equal segments at each iteration, as in standard binary search, we take a more conservative approach by splitting the interval using the cost ratio between the high-cost and low-cost queries. This splitting strategy minimizes the expected cost, rather than simply minimizing the expected number of queries. We refer to this procedure as \textbf{A}symmetric \textbf{S}earch (AS).\looseness=-1

\begin{figure}[t!] 
    \centering
    \vspace{-0.3cm}
    \begin{tabular}{@{}c@{}c}
        \includegraphics[height=0.17\textheight]{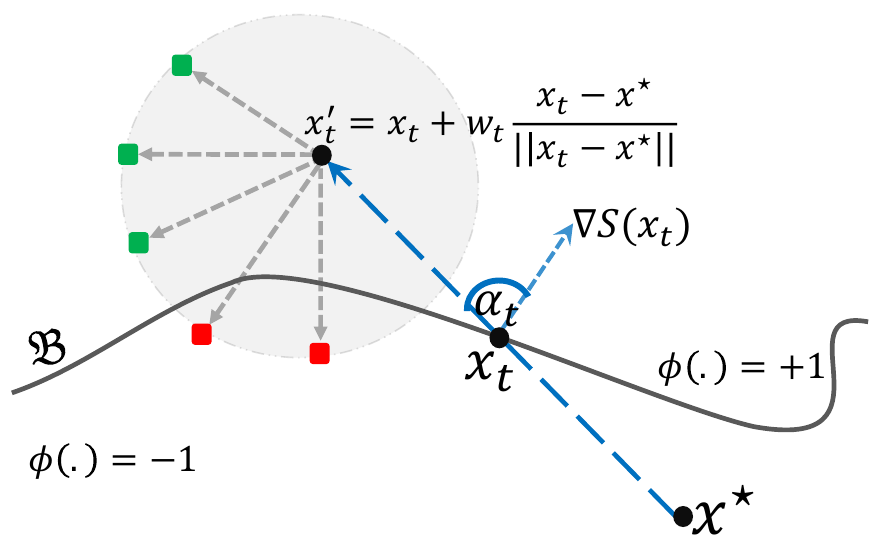} \hfill 
        \includegraphics[height=0.17\textheight]{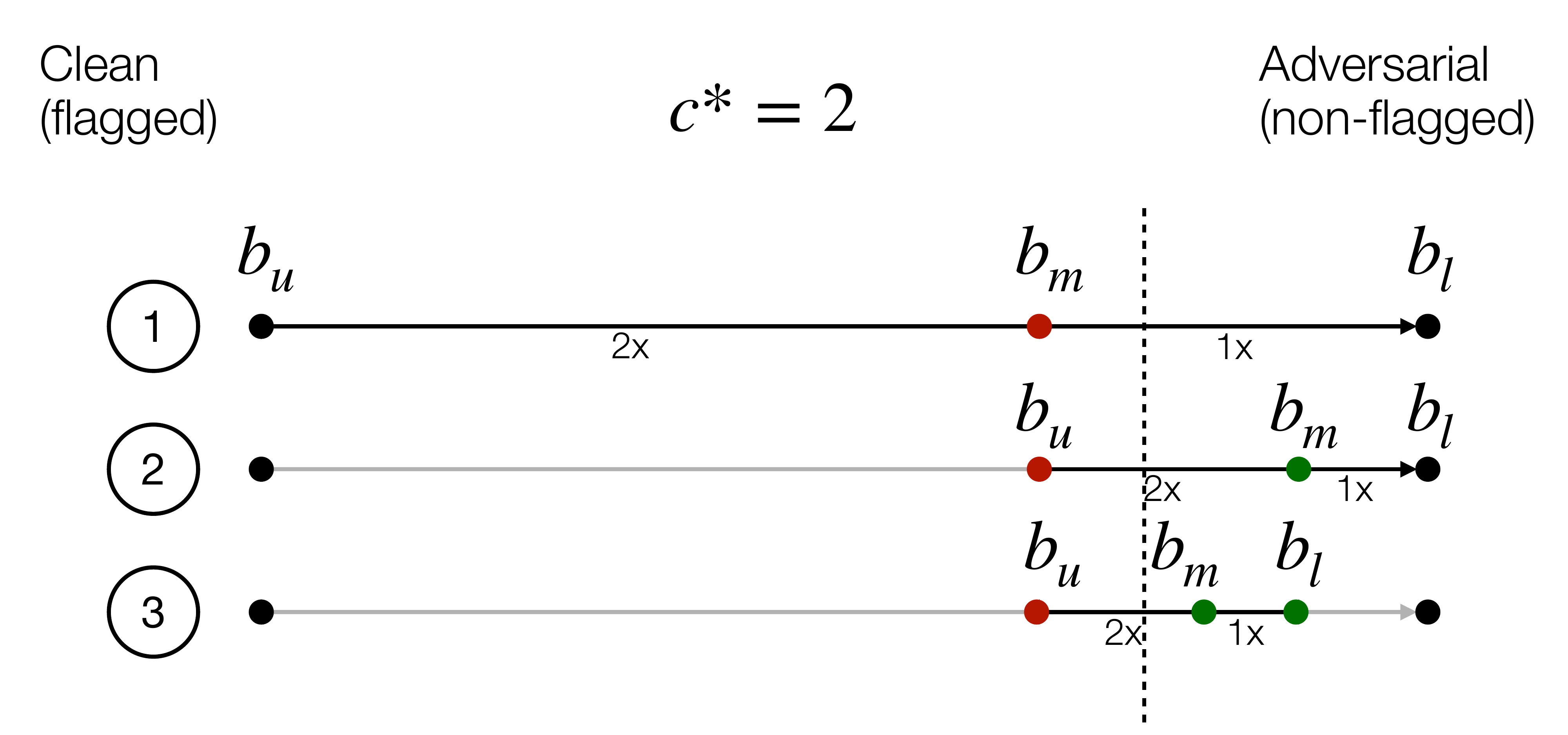} 
    \end{tabular}
    \vspace{-0.35cm}
    \caption{\textbf{Left.} Illustration of Asymmetric Gradient Estimation (AGREST), which reduces the frequency of high-cost queries by shifting the sampling region from $\mathbf{x}_t$ toward the adversarial region $\mathbf{x}^{\prime}_t$ and appropriately reweighting the outcomes.
    \textbf{{Right.}} Three steps of Asymmetric Search (AS) along the path from a clean (flagged) source image to an adversarial (non-flagged) image. Flagged queries are shown in red, non-flagged queries in green, and the dashed line denotes the decision boundary. \looseness=-1}
    \label{fig:Conceptual_Illustrations}
\end{figure}

Second, we adapt the gradient estimation procedure to account for arbitrary query costs. Instead of making queries around a norm ball centered at a boundary point, where approximately half of the queries are high-cost and half are low-cost as in standard HSJA, we shift the center to a point in the low-cost region and generate queries around it (\Cref{fig:Conceptual_Illustrations}). This adjustment intuitively reduces the frequency of high-cost queries, and the amount of shifting provides a natural way to control this frequency. To account for the shift, we weight high-cost and low-cost queries differently when estimating the gradient. We refer to this adaptation as \textbf{A}symmetric \textbf{GR}adient \textbf{EST}imation (AGREST). Our proposed framework is broadly compatible with a wide range of state-of-the-art decision-based attacks, including HSJA~\citep{chen2020hopskipjumpattack}, GeoDA~\citep{rahmati2020geoda}, CGBA~\citep{reza2023cgba}. It can be seamlessly applied to existing attack pipelines without requiring major structural changes. Through both theoretical analysis and rigorous experimental evaluation, we demonstrate that our method consistently outperforms existing attacks across arbitrary cost ratios. In particular, even under extreme asymmetry conditions where the cost of high-cost queries approaches infinity, our method incurs significantly less total query cost to achieve a given adversarial perturbation size compared to prior stealthy methods~\citep{debenedetti2024evading} (see Fig.~\ref{fig:comparison_stealthy}~(left)). This robustness highlights the effectiveness of our framework in balancing query efficiency and perturbation quality across diverse attack scenarios.\looseness=-1

The contribution of our paper is as follows:
\begin{itemize}
    \item To the best of our knowledge, we are the first to propose a versatile framework capable of handling arbitrary query cost ratios, providing flexibility across a wide range of adversarial attack scenarios.\looseness=-1
    
    \item Our framework introduces \textbf{AS} and \textbf{AGREST} as two core operations, optimizing the efficiency and effectiveness of adversarial attacks.\looseness=-1
    
    \item We provide a comprehensive theoretical analysis of the framework, establishing its foundations and demonstrating its robustness in diverse attack conditions.\looseness=-1
    
    \item We validate the framework through extensive empirical testing on benchmark datasets and models, including ImageNet, as well as advanced models such as CLIP and Vision Transformers and ResNet. This validation highlights the framework's superior performance.\looseness=-1
\end{itemize}

\section{Problem statement}
\label{sec:problem_statement}
\paragraph{An insight into unequal queries.}Consider an attacker trying to deceive an NSFW detector using decision-based methods. It may seem sufficient to choose an existing attack algorithm and add a small perturbation to an NSFW image based on that algorithm.
However, this approach may encounter some practical obstacles.
Most social networks enforce policies against uploading adult content, suspending users for violating these terms multiple times \citep{twitterSensitiveMedia}. 
Using the terminology from \citep{debenedetti2024evading}, this means that the cost of queries identified by the detector as NSFW, i.e., \textbf{flagged queries}, is higher than that of other queries, i.e., \textbf{non-flagged queries}.
For example, on $\mathbb{X}$, an attacker can make up to 2,400 posts per day on a single account \citep{twitterUnderstandingLimits}. However, after about 5 to 10 rule violations for uploading flagged posts, the attacker’s account will be suspended, requiring them to create a new one. On the other hand, in existing decision-based attacks, approximately half of the made queries are flagged \citep{debenedetti2024evading}. Therefore, if we assume the violation limit is 10, an attacker will be banned on $\mathbb{X}$ after about 20 posts. This example demonstrates the potential asymmetry in the costs of queries in a decision-based black-box setup.
\citep{debenedetti2024evading} addressed this asymmetry in costs by proposing stealthy attacks\footnote{\label{note1}Hereafter, we refer to prior stealthy attacks simply as \textit{stealthy attacks}, and to our approaches as \textit{asymmetric attacks} to emphasize their cost-aware design. Though inherently stealthy due to \textit{query cost awareness}, we adopt the term \textit{asymmetric} attacks to distinguish our method from prior work~\citep{debenedetti2024evading}.} designed to reduce the number of flagged queries. However, they overlooked the cost of non-flagged queries in their framework, leading to the generation of millions of non-flagged queries for every hundred flagged queries in stealthy attacks, which can also be costly.
For example, in the NSFW detector scenario, assume the attacker must create a new account after reaching the daily post limit. In stealthy attacks like HSJA, the attack can generate around $10^6$ non-flagged queries for every 100 flagged queries (\Cref{fig:comparison_stealthy}). While 100 flagged queries may lead to the creation of 10 new accounts, those $10^6$ non-flagged queries result in approximately 400 new accounts. This shows that non-flagged queries, despite being lower-cost, have a greater overall impact. Therefore, it is essential to develop generalized decision-based attacks that can effectively manage asymmetric query costs, making full use of low-cost queries without relying heavily on expensive ones.
\vspace{-0.33cm}
\paragraph{General formulation.}Assume that $f: \mathbb{R}^d \to \mathbb{R}^L$ is a pre-trained classifier with $L$ classes and parameters $\theta$. For an input image $\mathbf{x} \in \left[0, 1\right]^d$, $f_k\left(\mathbf{x}\right)$, the $k^{th}$ component of $f\left(\mathbf{x}\right)$, represents the predicted probability of the $k^{th}$ class. Additionally, for each correctly classified image $\mathbf{x}$ and query image $\mathbf{x}^\prime$, we define:
\begin{equation*}
\begin{aligned}
    S_{\mathbf{x}}\left(\mathbf{x}^\prime\right) = \underset{k\neq\hat{y}\left(\mathbf{x}\right)}{\argmax}\,f_k\left(\mathbf{x}^\prime\right) - f_{\hat{y}\left(\mathbf{x}\right)}\left(\mathbf{x}^\prime\right)\quad \text{and} \quad
    \phi_{\mathbf{x}}\left(\mathbf{x}^\prime\right) = \sign\left(S_{\mathbf{x}}\left(\mathbf{x}^\prime\right)\right)
\end{aligned}
\end{equation*}
Given a correctly classified source image $\mathbf{x}^\star$, the attacker's goal is to find the closest perturbed image $\mathbf{x}^\prime$ to the source image $\mathbf{x}^\star$ such that $\phi_{\mathbf{x}^\star}(\mathbf{x}^\prime) = 1$:
\begin{equation}
\label{eqn:adv_opt}
\underset{\mathbf{x}^\prime}{\minimize} \, \left\|\mathbf{x}^\star - \mathbf{x}^\prime\right\| \quad \text{s.t.} \quad \phi_{\mathbf{x}^\star}(\mathbf{x}^\prime) = 1
\end{equation}
Note that in a decision-based black-box setup, the attacker has access to $\phi_{\mathbf{x}^\star}(\mathbf{x}^\prime)$ but not $S_{\mathbf{x}^\star}(\mathbf{x}^\prime)$.
Previous methods sought to solve \Cref{eqn:adv_opt} while keeping the total number of queries as low as possible. However, as discussed, asymmetric query costs can make this approach ineffective. 
Instead, we have to keep the total cost of queries in an asymmetric setup as low as possible:
\begin{equation*}
\text{total cost} = N_{\text{source-class}} \cdot c^* + N_{\text{non-source-class}} \cdot 1 
\end{equation*}
Where $N_{\text{source-class}}$ for an attack is the number of queries $\mathbf{x}^\prime$ for which $\phi\left(\mathbf{x}^\prime\right) = -1$, $c^\star$ is the cost associated with this type of query, and $N_{\text{non-source-class}}$ is the number of queries $\mathbf{x}^\prime$ with a cost of 1 for which $\phi\left(\mathbf{x}^\prime\right) = 1$.
Existing decision-based attacks assume $c^\star=1$, while stealthy attacks assume $c^\star=\infty$. Our goal in this paper is to propose a framework that is effective for any arbitrary value of $c^\star$, unlike vanilla and stealthy attacks. In addition, we demonstrate that our approach outperforms stealthy attacks, even when $c^\star = \infty$.\looseness=-1

For brevity, in this paper, we omit $\mathbf{x}^\star$ when mentioning $S$ and $\phi$. Furthermore, we assume $c^\star \geq 1$, as our proposed framework is almost the same for $c^\star<1$. 
This allows us to refer to queries $\mathbf{x}^\prime$ where $\phi(\mathbf{x}^\prime) = -1$ as \textbf{high-cost queries} and others as \textbf{low-cost queries}.
These new terms reflect the concept of general asymmetric costs better than the previous terms used by \citep{debenedetti2024evading}, i.e., flagged and non-flagged queries.\looseness=-1
\vspace{-0.4cm}
\section{Proposed method}
\vspace{-0.3cm}
Decision-based black-box attacks typically involve two core operations, often applied iteratively to find small adversarial perturbations: 1. choosing a path, either straight, like GeoDA and HSJA \citep{rahmati2020geoda,chen2020hopskipjumpattack}, or curved, like SurFree and CGBA \citep{maho2021surfree,reza2023cgba}, and then 2. Searching along this path to find a new adversarial example, $\mathbf{x}_{t+1}$, that is closer to $\mathbf{x}^\star$ than $\mathbf{x}_t$, the adversarial example from the previous iteration.
These attacks either choose a path randomly, as in Boundary Attack \citep{brendel2017decision} and SurFree, or use queries to find a path that leads to a closer adversarial example than a random path, as in HSJA, GeoDA, and CGBA. 
To find this better-than-random path, these attacks estimate the normalized gradient direction of $S$ at $\mathbf{x}_t$ by approximating $\nabla S\left(\mathbf{x}_t\right)$ as follows:
\begin{equation}
\label{eqn:grad_est}
\widetilde{\nabla S}\left(\mathbf{x}_t\right) = \frac{1}{n_t} \sum_{i=1}^{n_t} \phi\left(\mathbf{x}_t + \delta \mathbf{u}_i\right) \mathbf{u}_i
\end{equation}
where $\delta$ is a small positive parameter and  $\mathbf{u}_1,\dots,\mathbf{u}_{n_t}$ are i.i.d. draws from either the uniform distribution over $\mathbb{S}^{d-1}$, the 
$\left(d-1\right)$-dimensional unit sphere, or the multivariate normal distribution. 
After finding a path, most attacks use variations of binary search to find $\mathbf{x}_{t+1}$. This is generally achieved by finding a boundary point $\mathbf{x}_{t+1}$ along the selected path, where $S\left(\mathbf{x}_{t+1}\right) = 0$, using binary search.\looseness=-1

As highlighted by \cite{debenedetti2024evading}, the issue with binary search and \Cref{eqn:grad_est} is that, in an asymmetric setup where $c^\star > 1$, crafting adversarial examples using these operations becomes costly because approximately half of the generated queries are high-cost. This raises the question of whether we can alter the distribution of generated queries to reduce the number of high-cost queries while maintaining the effectiveness of these two operations as observed in vanilla attacks. As a solution, we propose \textbf{AS} and \textbf{AGREST} techniques in the following sections.\looseness=-1
\subsection{Asymmetric Search~(AS)}
We define \( T: [0, 1] \to \mathbb{R}^d \) as the function that parameterizes the search path. For example, when the path is a straight line between the source image \( \mathbf{x}^\star \) and an adversarial image \( \mathbf{\Tilde{x}} \)~\cite{rahmati2020geoda, chen2020hopskipjumpattack}, the parameterization is given by \( T(\theta) = \theta\,\mathbf{x}^\star + (1 - \theta)\,\mathbf{\Tilde{x}} \). Similarly, when the search path follows a circular arc~\cite{reza2023cgba, maho2021surfree} between \( \mathbf{x}^\star \) and \( \mathbf{\Tilde{x}} \), lying on a circle in the 2D subspace spanned by \( \mathbf{u} = \frac{\mathbf{\Tilde{x}} - \mathbf{x}^\star}{\left\|\mathbf{\Tilde{x}} - \mathbf{x}^\star\right\|_2} \) and a unit vector \( \mathbf{v} \) satisfying \( \langle \mathbf{u}, \mathbf{v} \rangle = 0 \), the parameterization is given by \( T(\theta) = \mathbf{x}^\star + \cos\left(\tfrac{\pi}{2} \theta\right) \cdot \left\|\mathbf{\Tilde{x}} - \mathbf{x}^\star\right\|_2 \left( \cos\left(\tfrac{\pi}{2} \theta\right) \mathbf{u} + \sin\left(\tfrac{\pi}{2} \theta\right) \mathbf{v} \right) \).\looseness=-1

\begin{wrapfigure}[17]{r}{0.45\textwidth} 
\begin{minipage}{0.45\textwidth}
\vspace{-0.73cm}
\begin{algorithm}[H]
    \caption{AS Algorithm}
    \centering
    \label{alg:as-algorithm}
    \begin{algorithmic}[1]
    \REQUIRE Parametrization function $T$, threshold $\tau$ 
    \ENSURE Near-boundary adversarial example
    \STATE $b_l \gets 0,\, b_u \gets \ceil*{\frac{1}{\tau}}$
    \WHILE {$b_u - b_l > 1$}
    \STATE $b_m \gets b_l + \ceil*{\frac{b_u - b_l}{\left(c^\star+1\right)}}$
    \IF{$\phi\left(T\left(b_m\tau\right)\right) = 1$}
        \STATE $b_l \gets b_m$
    \ELSE
        \STATE $b_u \gets b_m$
    \ENDIF
    \ENDWHILE
    \STATE \textbf{return} $T(b_l\tau)$
\end{algorithmic}
\end{algorithm}
\end{minipage}
\end{wrapfigure}
As mentioned, the objective is to find the along-the-path boundary point $T\left(\theta^\star\right)$ for some $\theta^\star \in [0, 1]$.
However, since the search space is discrete in practice, our goal is to find a point $\mathbf{x}_{t+1}$ that is near the boundary, i.e., where $S\left(\mathbf{x}_{t+1}\right) \approx 0$.
This implies that for a given error threshold $\tau$, we need to find $[a,b] \subset [0,1]$ such that $0 < b-a \leq \tau$ and $\theta^\star \in [a, b]$. Since $S$ is continuous, one approach to achieve this is to find $0 \leq k \leq \lceil \frac{1}{\tau} \rceil$ for which $\phi\left(T\left(k\tau\right)\right) = 1$ and $\phi\left(T\left((k+1)\tau\right)\right) = -1$.
The smaller $\tau$ is, the closer we get to the boundary, although this requires more queries.\looseness=-1 

It is well-known that in one-dimensional search, binary search is the optimal comparison-based algorithm in terms of minimizing the expected number of queries, assuming that the boundary point is uniformly distributed along the path (see Assumption~\ref{as:search_as}).
Nonetheless, in the asymmetric cost setting, the expected cost of binary search is $\Theta\left(c^\star \log\left(1/\tau\right)\right)$, because the expected number of queries is $\Theta\left(\log\left(1/\tau\right)\right)$, and about half of these queries are expected to incur the higher cost $c^\star$.\looseness=-1
\begin{assumption}
\label{as:search_as}
Let $\Theta^\star_{rv} \in [0, 1)$ be a random variable. If $T(\Theta^\star_{rv})$ lies on the decision boundary, that is, $S(T(\Theta^\star_{rv})) = 0$, then we assume $\Theta^\star_{rv}$ is drawn uniformly from $[0, 1)$.
\end{assumption}
The core idea behind AS is similar to that of binary search, but with a more conservative strategy to account for asymmetric costs.
Instead of splitting the interval into two equal parts, AS divides it with a $1:c^\star$ ratio at each step, favoring lower-cost queries.
More specifically, suppose we know the desired point lies within $\left[b_l \tau, b_u \tau\right] \subset \left[0, 1\right]$.
Then, as shown in \Cref{alg:as-algorithm}, if $\phi\left(T\left(b_l \tau + \left\lceil \frac{b_u - b_l}{c^\star + 1} \right\rceil \tau\right)\right) = 1$, AS continues the search in $\left[b_l \tau + \left\lceil \frac{b_u - b_l}{c^\star + 1} \right\rceil \tau, b_u \tau\right]$; otherwise, it proceeds within $\left[b_l \tau, b_l \tau + \left\lceil \frac{b_u - b_l}{c^\star + 1} \right\rceil \tau\right]$.
This process is repeated until AS locates a point within $\tau$ of the boundary.\looseness=-1

Note that when $c^\star = 1$, AS reduces to standard binary search, and when $c^\star = \infty$, it becomes a simple line search strategy, as used in stealthy attacks~\citep{debenedetti2024evading}, where the algorithm checks $\tau, 2\tau, 3\tau, \dots$ sequentially.
The expected cost of AS is given in \Cref{the:as_cost}, showing that it improves over binary search by a factor of $\Theta\left(\log\left(c^\star+1\right)\right)$.\looseness=-1
\begin{theorem}
\textbf{(Cost Analysis of AS)}
\label{the:as_cost}
Suppose $0 < \tau < 1$ and $c^\star \geq 1$. Under \Cref{as:search_as}, the expected cost of the AS algorithm is $\mathcal{O}(c^\star \log_{(c^\star+1)} \left(1/\tau\right))$.\looseness=-1
\end{theorem}
To illustrate the effect of AS in practice, we compare the cost of AS and binary search when \( c^\star = 10^3 \), and we observe that the cost of binary search is approximately 2.5 times higher than that of AS. The results are provided in \Cref{app:emp_study}. An illustration of the AS algorithm can be found in~\Cref{app:Conceptual_Illustration}~(Right) and~\cref{fig:Conceptual_Illustrations_app}.\looseness=-1
\vspace{-0.3cm}
\subsection{\textbf{A}symmetric \textbf{Gr}adient \textbf{Est}imation (AGREST)}
\vspace{-0.26cm}
As mentioned earlier, to keep the gradient estimation (i.e., \Cref{eqn:grad_est}) effective in the asymmetric setup, our goal is to adjust the distribution of queries generated during the process. To achieve this, we first propose a family of estimations that provides flexibility in adjusting the distribution of queries used for estimation. Then, we introduce a method to select the estimator within this family that maximizes the similarity between the approximated gradient and the true gradient.\looseness=-1
\vspace{-0.3cm}
\paragraph{How can we control the query distribution?}
The main idea behind AGREST is to estimate $\nabla S\left(\mathbf{x}_t\right)$ by approximating the gradient at a point like $\mathbf{x}^\prime_t$, which is close to $\mathbf{x}_t$. This approach allows us to alter the distribution of made queries while approximating effective directions for the attacks.
In this method, we move $\mathbf{x}_t$ away from $\mathbf{x}^\star$ by $\omega_t$, the overshooting value, to reach the new point $\mathbf{x}^\prime_t = \mathbf{x}_t + \omega_t \frac{\mathbf{x}_t-\mathbf{x^\star}}{\left\|\mathbf{x}_t-\mathbf{x^\star}\right\|_2}$(\Cref{fig:Conceptual_Illustrations}~(left)).
Then, we use \Cref{eqn:grad_est} at $\mathbf{x}^\prime_t$ almost similar to the vanilla estimation, except that AGREST assigns more weight to high-cost queries than to low-cost ones (using importance sampling).
In other words, we estimate $\nabla S$ as follows:
\begin{equation*}
\begin{split}
    \widehat{\nabla S}\left(\mathbf{x}_t,\omega_t,\beta_t\right) &= \frac{1}{n_t} \sum_{i=1}^{n_t} \widehat{\phi}_t\left(\mathbf{x}^\prime_t + \delta \mathbf{u}_i\right) \mathbf{u}_i 
    = \frac{1}{n_t} \sum_{i=1}^{n_t} \widehat{\phi}_t\left(\mathbf{x}_t + \omega_t \frac{\mathbf{x}_t-\mathbf{x^\star}}{\left\|\mathbf{x}_t-\mathbf{x^\star}\right\|_2} + \delta \mathbf{u}_i\right) \mathbf{u}_i
\end{split}
\end{equation*}
where \( \widehat{\phi}_t(\mathbf{x}) = (1 - \beta_t)\, \mathbf{1}\{\phi(\mathbf{x}) = 1\} - \beta_t\, \mathbf{1}\{\phi(\mathbf{x}) = -1\} \) is the sampling weight function, \( \frac{1}{2} \leq \beta_t < 1 \) is the sampling weight parameter, \( \mathbf{u}_1, \dots, \mathbf{u}_{n_t} \) are i.i.d. draws from \textsc{Uniform}\( (\mathbb{S}^{d-1}) \), and \( \mathbf{1}\{\cdot\} \) denotes the indicator function. The parameters \( \omega_t \) and \( \beta_t \) allow us to control the likelihood of making high-cost queries and their associated weight in our estimation.\looseness=-1
\paragraph{How can we choose the best AGREST estimator?}
To ensure the selected estimation is as close as possible to the true gradient direction among all AGREST estimators, one potential solution is to find the estimator that maximizes:
\begin{equation*}
    \mu(\mathbf{x}_t, \omega_t, \beta_t, n_t)=\mathbb{E}_{\mathbf{u}_{1:n_t}} \cos\left(\nabla S\left(\mathbf{x}_t\right), \widehat{\nabla S}\left(\mathbf{x}_t, \omega_t, \beta_t\right)\right)
\end{equation*}
within the query budget, where $\cos$ represents the cosine similarity function.
To calculate this function, we need to assume that $S$
has certain characteristics. One common choice is to assume that $S$ is $L$-smooth. However, this assumption introduces excessive complexity to our analysis and may add additional hyperparameters related to $L$ to the current set of hyperparameters in the existing attacks. Therefore, similar to \cite{rahmati2020geoda, maho2021surfree} and based on observations from \cite{fawzi2016robustness, fawzi2017robustness, fawzi2018empirical}, we assume that $S$ is locally linear around $\mathbf{x}_t$, $S(\mathbf{x}_t^\prime+\delta\mathbf{u})\approx S(\mathbf{x}_t) + \langle\nabla S(\mathbf{x}_t), \mathbf{x}_t^\prime + \delta\mathbf{u}-\mathbf{x}_t\rangle$.
Since $\mathbf{x}_t$ is a boundary point, $S(\mathbf{x}_t)=0$. Thus, we have:
\begin{equation}
\begin{split}
    \label{eqn:linear_as}
    \phi(\mathbf{x}_t^\prime+\delta\mathbf{u}) &\approx \sign\left(\langle\nabla S(\mathbf{x}_t), \mathbf{x}_t^\prime + \delta\mathbf{u}-\mathbf{x}_t\rangle\right)
        =\sign(\cos\alpha_t\cdot\omega_t + \langle\mathbf{g}_t,\delta\mathbf{u}\rangle)
\end{split}
\end{equation}
where $\mathbf{g}_t = \frac{\nabla S(\mathbf{x}_t)}{\left\|\nabla S(\mathbf{x}_t)\right\|_2}$ and $\alpha_t$ is the angle between $\mathbf{x}_t-\mathbf{x}^\star$ and $\mathbf{g}_t$ (\Cref{fig:Conceptual_Illustrations}). 
Additionally, based on this assumption, we can calculate the probability of low-cost queries, namely $p_t(\omega_t) = \mathbb{P}[\phi(\mathbf{x}^\prime_t + \delta\mathbf{u}) = 1]$, using \Cref{lem:chudnov}. 
\begin{lemma}
\label{lem:chudnov}
\textbf{(Hyperspherical Cap~\cite{chudnov1986minimax})} 
Under local linearity around $\mathbf{x}_t$ (\Cref{eqn:linear_as}), we have
$p_t(\omega_t) = \frac{1}{2}\left(1 + \mathcal{I}_{d-2}(\delta^{-1}\cos\alpha_t \omega_t) / \mathcal{I}_{d-2}(0)\right)$, where $\mathcal{I}_{d}(x) = \int_0^{1-x} (1 - t^2)^{(d-1)/2} dt$.
\end{lemma}
Based on \Cref{lem:chudnov}, we can infer that $p_t(\omega_t)$ is strictly increasing and therefore invertible. Nonetheless, even with the linearity assumption, calculating the expected value remains challenging due to the nonlinearity of cosine similarity and the complexity of handling multiple independent random vectors. Therefore, inspired by measure concentration~\cite{ledoux2001concentration}, we approximate $\mu(\mathbf{x}_t, \omega_t, \beta_t, n_t)$ as follows:
\begin{equation*}
J(\mathbf{x}_t, \omega_t, \beta_t, n_t) =
\left( n_t^{1/2} \cdot \mathbb{E}\left[\widehat{\phi}(\mathbf{x}_t'+\delta\mathbf{u})\langle\mathbf{g}_t, \mathbf{u}\rangle \right] \right)
\cdot \left( \mathbb{E}\left[\widehat{\phi}(\mathbf{x}_t'+\delta\mathbf{u})^2 \right] \right)^{-1/2}
\end{equation*}
This new objective is easier to calculate since it removes the need to deal with multiple random vectors. \Cref{the:age_obj} establishes a convergence bound for the approximation.
\begin{theorem}
\textbf{(Expected Cosine Similarity Approximation)} 
    \label{the:age_obj}
    Under the local linearity assumption around $\mathbf{x}_t$, for any constants $0< z < \frac{1}{8}$ and $\frac{1}{2} \leq q, \beta < 1$, as $n_t$ and $d$ approach infinity, we have
    \begin{equation*}
        \left|\frac{\mu(\mathbf{x}_t, p^{-1}_t(q), \beta, n_t)}{J(\mathbf{x}_t, p^{-1}_t(q), \beta, n_t)}-1\right|\leq \mathcal{O}\left(d^{-z}\right)
    \end{equation*}
\end{theorem}
Now, we can formulate the optimization problem. The goal is to maximize $J(\mathbf{x}_t, \omega_t, \beta_t, n_t)$ within a query budget. Specifically, we want to:
\begin{equation}
\begin{aligned}
\label{eqn:age_opt}
\max_{\omega_t, \beta_t, n_t}\quad J(\mathbf{x}_t, \omega_t, \beta_t, n_t) \qquad
\text{s.t.} \quad n_t(c^\star - (c^\star-1)p_t(\omega_t)) \leq c_t
\end{aligned}
\end{equation}
where $c_t$ is the maximum allowed cost of estimation at iteration $t$ of the algorithm. The constraint in \Cref{eqn:age_opt}
ensures that the expected estimation cost at iteration $t$ is at most $c_t$. To solve this optimization problem, we propose \Cref{the:sol_obj}.
\begin{theorem}
\textbf{(Optimal AGREST Parameters)} 
    \label{the:sol_obj}
    Suppose the solution to \Cref{eqn:age_opt} is represented by $(\omega_t^\star, w_t^\star, n_t^\star)$. Given the local linearity around $\mathbf{x}_t$, the following statements hold:
    \begin{enumerate}
        \item $n_t^\star = c_t \left( c^\star - (c^\star - 1)p_t(\omega_t^\star) \right)^{-1}$ and $\beta_t^\star = p_t(\omega_t^\star)$.
        \item $\omega_t^\star$ maximizes the following function over the interval $\left[0, \delta / \cos\alpha_t\right]$:
        \begin{equation*}
        \widehat{J}_t(\omega_t) =
\left( 1 - \left(\delta^{-1}\cos\alpha_t  \omega_t \right)^2 \right)^{d-1}
 \Big( p_t(\omega_t)  \left(1 - p_t(\omega_t)\right) \left( c^\star - (c^\star - 1)p_t(\omega_t) \right) \Big)^{-1}
        \end{equation*}
    \end{enumerate}
\end{theorem}
An immediate consequence of \Cref{lem:chudnov} and \Cref{the:sol_obj} is the existence of $\omega^\star$ such that $\omega^\star = \cos\alpha_1 \cdot \omega_1^\star = \cos\alpha_2 \cdot \omega_2^\star = \dots$. As a result, we aim to find $\omega^\star$. One problem is that $\mathcal{I}_d$ has a complex closed form. Thus, finding a closed form for $\omega^\star$ would be challenging. Instead, we use numerical methods to evaluate this integral and numerical optimization techniques to find $\omega^\star$. 
In particular, we use QUADPACK \cite{piessens2012quadpack} for the integral calculation and the Nelder–Mead method \cite{nelder1965simplex} for maximizing $\widehat{J}_t(\omega_t)$.
Another problem is that we need to know $\alpha_t$ in order to obtain the optimal overshooting value, which is not possible in a black-box setup. Thus, we need to estimate $\alpha_t$ at each iteration $t$. 
\paragraph{How can we estimate $\alpha_t$?} 
\begin{wrapfigure}[10]{r}{0.45\textwidth}
\begin{minipage}{0.45\textwidth}
\vspace{-0.8cm}
\begin{algorithm}[H]
    \caption{Overshooting Scheduler Step}
    \centering
    \label{alg:scheduler-algorithm}
    \begin{algorithmic}[1]
    \REQUIRE Iteration $t$, dimension $d$, desired probability $p$, scheduler rate $m$
    \ENSURE Next cosine value $\alpha_{t+1}$
    \STATE $\alpha_1 \gets \textsc{Init-Angle}\left(d\right)$ \hfill $\triangleright$ ~\Cref{the:cos_ini}
    \STATE $\Hat{\alpha}_{t+1} \gets 1 - \left(1-\cos\alpha_1\right) \left(t+1\right) ^ {-m}$
    \STATE $\alpha_{t+1} \gets \arccos \left(\Hat{\alpha}_{t+1}\right)$
    \RETURN $\alpha_{t+1}$
\end{algorithmic}
\end{algorithm}
\end{minipage}
\end{wrapfigure}
We split this problem into two steps: 1. estimating $\alpha_1$, and 2. understanding the behavior of $\alpha_t$ with respect to $t$.
For the first problem, initially, we expect $\mathbf{x}_1 - \mathbf{x}^\star$ and $\mathbf{g}_1$ to be somehow independent, as most existing attacks select $\mathbf{x}_1$ using a random direction. The only reasonable assumption about these two vectors is that they likely have a positive correlation, i.e., $\langle\mathbf{x}_1 - \mathbf{x}^\star, \mathbf{g}_1\rangle\geq 0$. 
Specifically, if we know that $\mathbf{x}_1$ is the closest boundary point to $\mathbf{x}^\star$ along the direction of $\mathbf{x}_1-\mathbf{x}^\star$, meaning there is no $0< r < 1$ such that $\phi(\mathbf{x}^\star + r(\mathbf{x}_1-\mathbf{x}^\star)) = 1$, this assumption provably holds. Given this assumption, we can use \Cref{the:cos_ini} to attain $\alpha_1$.\looseness=-1 
\begin{theorem}
\textbf{(Initial Cosine Value)} 
    \label{the:cos_ini}
    Under local linearity around $\mathbf{x}_1$, if there is no $0 < r < 1$ such that $\phi(\mathbf{x}^\star + r(\mathbf{x}_1 - \mathbf{x}^\star)) = 1$, then we have 
    $\mathbb{E}\left[\cos\alpha_1\right] = \Gamma\left(\tfrac{d}{2}\right) \left( 2\sqrt{\pi}\,\Gamma\left(\tfrac{d+1}{2}\right) \right)^{-1}$.
\end{theorem}
\begin{algorithm}
    \caption{AGREST Estimation}
    \centering
    \label{alg:age-algorithm}
    \begin{algorithmic}[1]
    \REQUIRE Iteration $t$, source image $\mathbf{x}^\star$, boundary point $\mathbf{x}_t$, dimension $d$, high-cost query cost $c^\star$, sampling radius $\delta$, sampling batch size $b$, cosine value $\alpha_t$, vanilla gradient estimation query budget $n^\prime_t$, scheduler rate $m$
    \ENSURE Normalized approximated direction $g_t$, next cosine value $\alpha_{t+1}$
    \STATE $n_L \gets 0,\enspace n_H \gets 0, \enspace \mathbf{v}^+ \gets \vec{0},\enspace \mathbf{v}^- \gets \vec{0},\enspace \hat{c} \gets 0,\enspace \omega^\star \gets \textsc{Overshooting}\left(c^\star\right)$ \hfill $\triangleright$ \Cref{the:sol_obj}
    \STATE $\omega_t \gets \omega^\star/\cos\alpha_t, \enspace c_t \gets n^\prime_t(c^\star+1)/2$
    \WHILE {$\hat{c} < c_t$}
    \FOR{\textbf{each} $\mathbf{u}_i \sim \textsc{Uniform}(\mathbb{S}^{d-1})$, $i = 1, \ldots, b$}
    \IF{$\phi\left(\mathbf{x}_t + \omega_t \frac{\mathbf{x}_t-\mathbf{x^\star}}{\left\|\mathbf{x}_t-\mathbf{x^\star}\right\|_2} + \delta \mathbf{u}_i\right) = 1$}
        \STATE $\mathbf{v}^+ \gets \mathbf{v}^+ + \mathbf{u}_i,\enspace n_L \gets n_L + 1, \enspace \hat{c} \gets \hat{c} + 1$
    \ELSE
        \STATE $\mathbf{v}^- \gets \mathbf{v}^- - \mathbf{u}_i, \enspace n_H \gets n_H + 1, \enspace \hat{c} \gets \hat{c} + c^\star$
    \ENDIF 
    \ENDFOR
    \ENDWHILE
    \STATE $\hat{p}_t \gets n_L / \left(n_L + n_H\right)$
    \STATE $g_t \gets \left(1-\hat{p}_t\right)\mathbf{v}^+ + \hat{p}_t\mathbf{v}^-,\enspace\alpha_{t+1}\gets\textsc{Scheduler-Step}\left(t, \hat{p}_t, m\right)$ \hfill $\triangleright$ \Cref{alg:scheduler-algorithm}
    \RETURN $g_t / \left\|g_t\right\|_2,\enspace\alpha_{t+1}$
\end{algorithmic}
\end{algorithm}
The next step is to estimate $\alpha_t$ after the first iteration. Chen et al.~\cite{chen2020hopskipjumpattack} showed that in HSJA, $\cos(\mathbf{x}_t - \mathbf{x}^\star, \mathbf{g}_t) \geq 1 - c\,t^{-m}$ for some constant $c$ and $0 < m < \frac{1}{2}$.
 This motivated us to heuristically estimate $\alpha_t$ as $\arccos (1 - (1-\cos\alpha_1) t ^ {-m})$, where $m$ is a new hyperparameter introduced to the existing attacks (\Cref{alg:scheduler-algorithm}).
As stated in \Cref{the:sol_obj}, under the assumption of local linearity, the value of $\beta_t$ in an optimal AGREST estimation is the probability of making low-cost queries. 
However, similar to \cite{chen2020hopskipjumpattack}, we use the empirical probability of making low-cost queries, namely $\frac{n_{\text{L}}}{n_{\text{L}} + n_{\text{H}}}$, to reduce the variance of the estimation. Here, $n_{\text{H}}$ and $n_{\text{L}}$ represent the number of high-cost and low-cost queries made in an AGREST estimator, respectively.
Additionally, we set $c_t$ to the expected cost of the vanilla attack, namely $\frac{n'_t(c^\star + 1)}{2}$, where $n'_t$ is the number of made queries by the vanilla attack at iteration $t$. A detailed overview of AGREST is provided in \Cref{alg:age-algorithm}.
Note that in practice, most attacks are performed in a given subspace rather than in the entire space to improve sample efficiency. In these cases, we use the effective dimension \( d' \) of the subspace instead of \( d \), the dimension of the original space. For more details, see \Cref{app:imp_dim_red}.
Finally, it is worth mentioning that the probability of making low-cost queries in AGREST closely follows our theoretical analysis in practice. Further details and empirical results are presented in \Cref{app:emp_study}. \looseness=-1
\section{Experiments}
\begin{figure*}[h!] 
    \centering
    \vspace{-0.3cm}
    \begin{tabular}{@{}c@{}c@{}c}
        \includegraphics[width=0.33\textwidth]{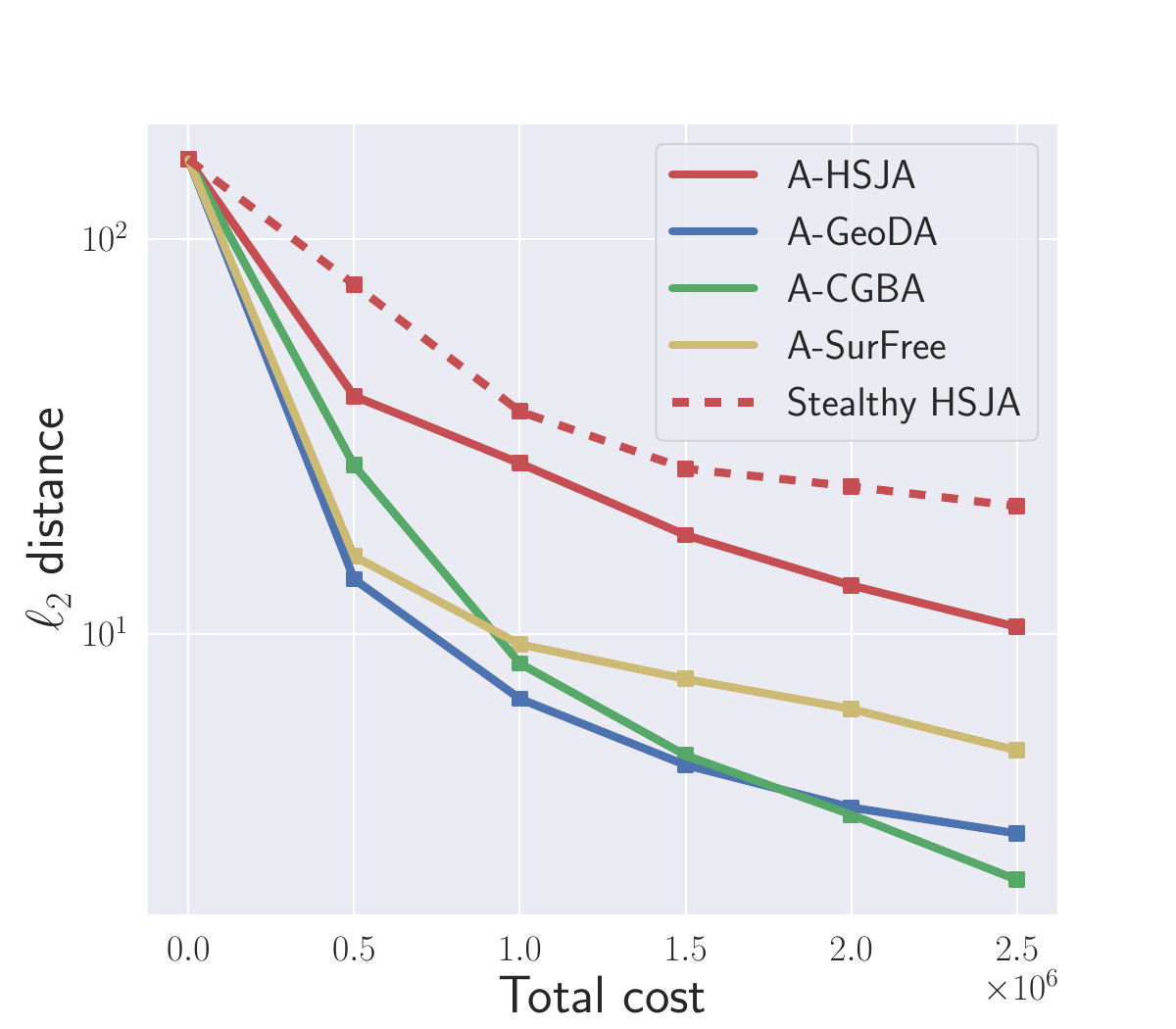} & 
        \includegraphics[width=0.33\textwidth]{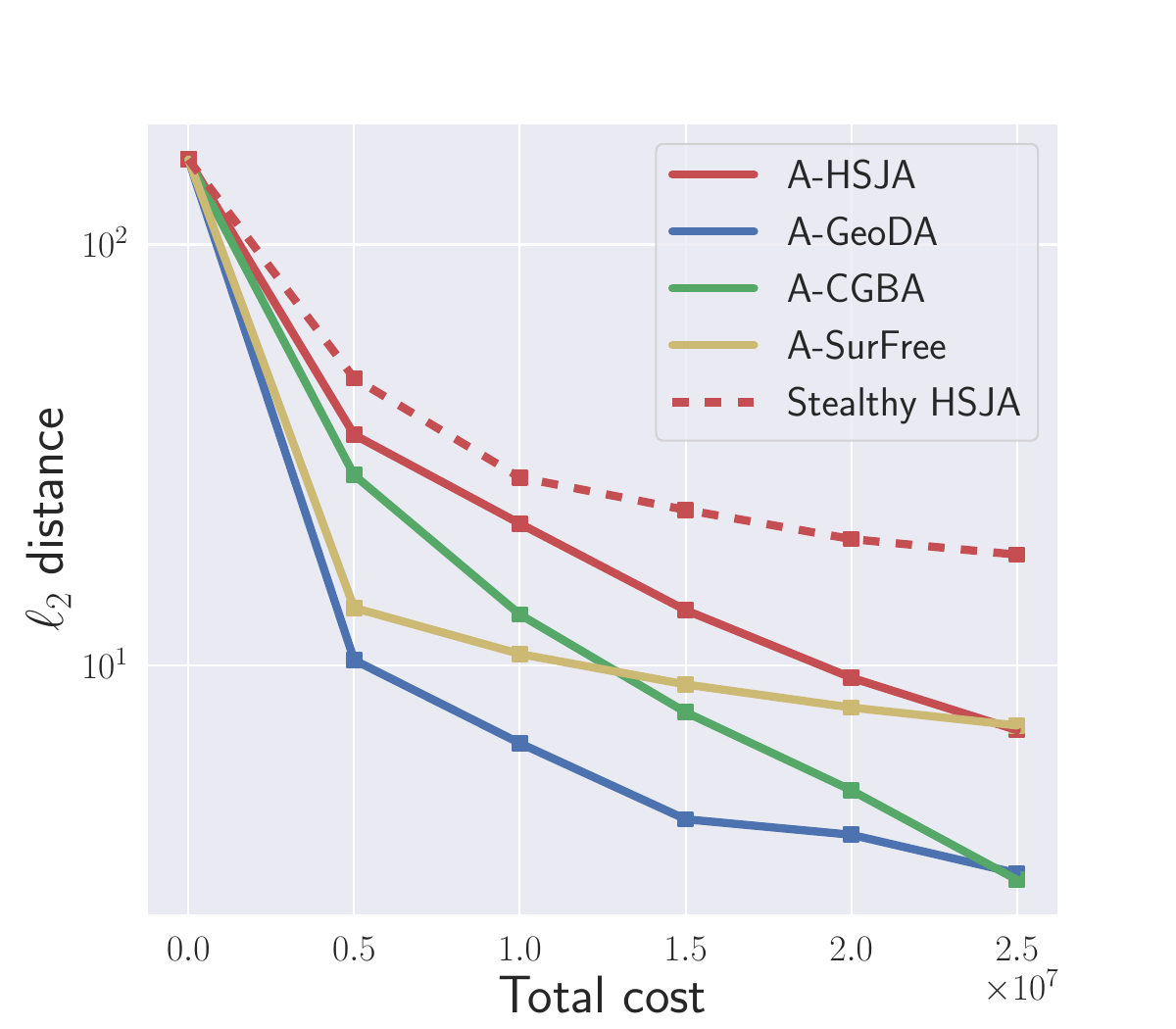} & 
        \includegraphics[width=0.33\textwidth]{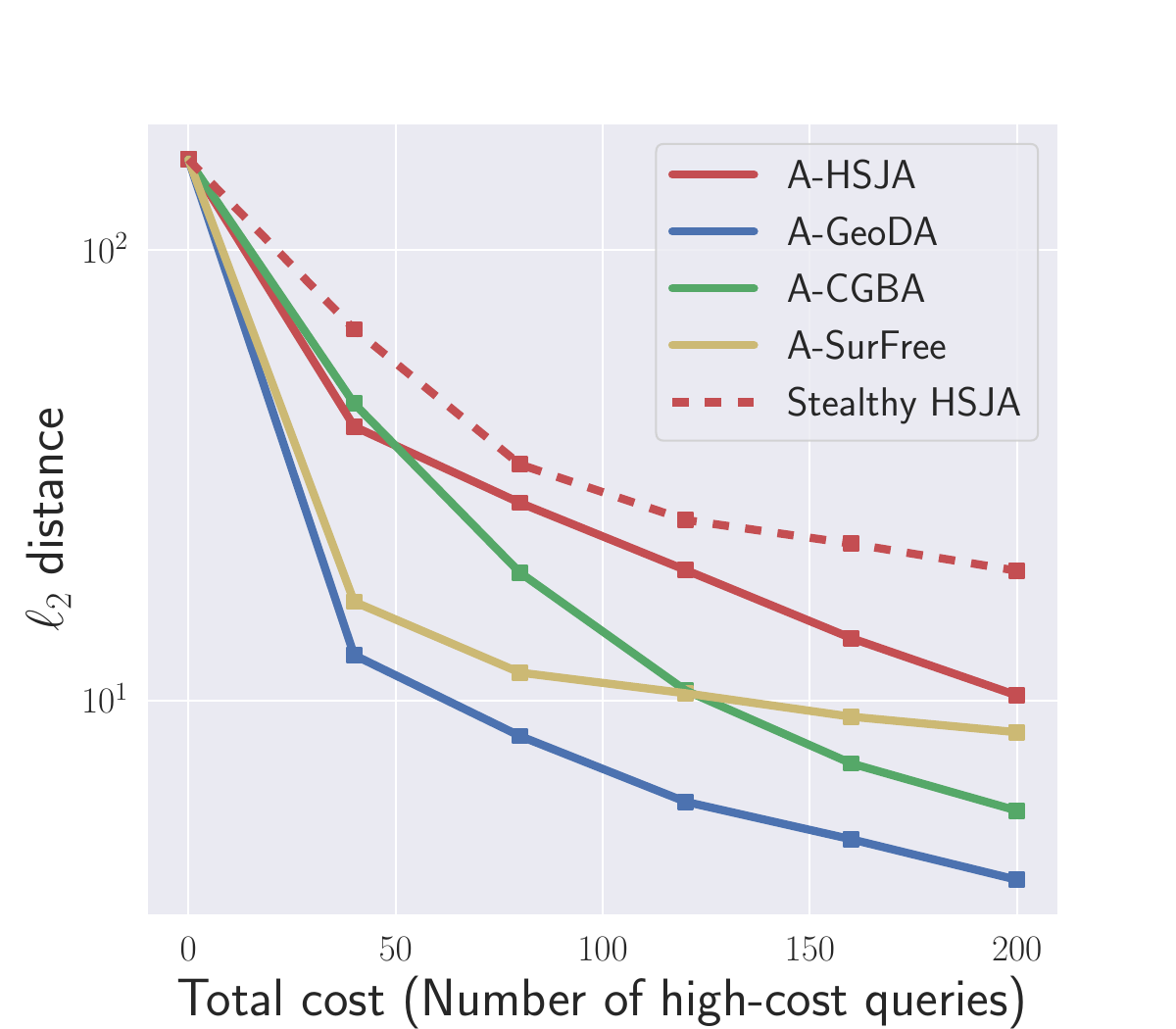}
    \end{tabular}
    \vspace{-0.35cm}
    \caption{\textbf{Performance of various asymmetric attacks compared to Stealthy HSJA under high cost asymmetry with ResNet-50.}
    The value of \( c^\star \) is \( 10^4 \), \( 10^5 \), and \( \infty \) from left to right.}
    \label{fig:stealthy_aa}
\end{figure*}

\vspace{-0.3cm}
\paragraph{Model, dataset, and metric.} For robust evaluation of asymmetric attacks, we employed ImageNet-trained models: ResNet-50~\citep{he2016deep}, ViT-B/32, ViT-B/16~\citep{dosovitskiy2021imageworth16x16words}, and CLIP~\citep{radford2021learningtransferablevisualmodels}. Original images~($\mathbf{x^\star}$) were 500 correctly classified ImageNet validation samples similar to the~\cite{debenedetti2024evading}. Numerical tasks used SciPy~\citep{virtanen2020scipy}. Attack performance was measured by the median $\ell_2$ distance between perturbations and originals over query costs, consistent with previous work.\looseness=-1
\vspace{-0.35cm}
\paragraph{Attacks and hyperparameters.}
To evaluate our proposed framework, we modify SurFree, HSJA, GeoDA, and CGBA by changing their respective search methods to AS and, where applicable, updating their gradient estimation to asymmetric gradient estimation. The other components of the attacks remain similar to their original implementations, except for the line search along the gradient direction in GeoDA. We slightly modify this part to reduce the number of high-cost queries generated during this phase (see \Cref{app:imp}). 
Moreover, to compare our framework with stealthy attacks, we use Stealthy HSJA, as it outperforms other stealthy attacks when the $\ell_2$ norm is the evaluation metric (Fig. 3 of \cite{debenedetti2024evading}).
For the hyperparameters used in the vanilla attacks, we generally use the same values. The only exception is the subspace method in SurFree. Specifically, instead of using the $\text{DCT}_{8 \times 8}$ method in SurFree, we set it to $\text{DCT}_{\text{full}}$. This adjustment allows for a fair comparison of SurFree with other attacks, as GeoDA and CGBA both use the $\text{DCT}_{\text{full}}$ technique. 
Furthermore, we set the newly introduced hyperparameter $m$, the overshooting scheduler rate, to $0.02$, $0.06$, and $0.06$ for HSJA, GeoDA, and CGBA, respectively (for more details on the effect of hyperparameter $m$ see \Cref{app:imp_hyperparam}).\looseness=-1
\begin{table*}[t!]
  \caption{Median \( \ell_2 \) distance for various \( c^\star \) values and different types of attacks across neural network architectures. VA stands for Vanilla Attack. The bold numbers represent the best performance among different variants of each attack for each $c^\star$ value and model~(For a comprehensive analysis of attacks under varying total cost constraints, we refer readers to \Cref{t:results_table_ViT_B_32} and \Cref{t:results_table_ViT_B_16} in \Cref{app:ViT_additional_results}, which present exhaustive experimental results across different total cost budgets and query cost $c^\star$.)\looseness=-1}
  \vspace{-0.28cm}
  \label{t:merged-results}
  \centering
  \resizebox{1\textwidth}{!}{
  \begin{tabular}{ccccccccccc}
    \toprule
    & & \multicolumn{4}{c}{ResNet-50} & \multicolumn{4}{c}{ViT-B/32} \\
    \cmidrule(lr){3-6} \cmidrule(lr){7-10}
    & & \cellcolor{gray!8}$c^\star=2$ & \cellcolor{gray!8}$c^\star=5$ & \cellcolor{gray!8}$c^\star=10^2$ & \cellcolor{gray!8}$c^\star=10^3$ & \cellcolor{gray!8}$c^\star=2$ & \cellcolor{gray!8}$c^\star=5$ & \cellcolor{gray!8}$c^\star=10^2$ & \cellcolor{gray!8}$c^\star=10^3$ \\
    \midrule
    \multirow{2}{*}{\textit{Attack}} & \multirow{2}{*}{\textit{Method}} & \multicolumn{8}{c}{\textit{Total Cost of Queries}} \\
    \cmidrule(lr){3-10}
    & & \cellcolor{gray!8}10K & \cellcolor{gray!8}15K & \cellcolor{gray!8}150K & \cellcolor{gray!8}250K & \cellcolor{gray!8}10K & \cellcolor{gray!8}15K & \cellcolor{gray!8}150K & \cellcolor{gray!8}250K \\
    \midrule
    \multirow{2}{*}{SurFree} 
    & VA & 4.09 & 5.19 & 5.21 & 17.49 & 2.9 & 2.5 & 5.13 & 16.12 \\
    & VA+AS (A-SurFree) & \textbf{3.45} & \textbf{3.52} & \textbf{3.80} & \textbf{7.59} & \textbf{2.4} & \textbf{2.1} & \textbf{3.68} & \textbf{6.35} \\
    \midrule
    \multirow{4}{*}{HSJA} 
    & VA & 2.24 & 2.77 & 4.66 & 23.72 & 18.3 & 13.9 & 4.21 & 22.46 \\
    & VA+AS & 2.16 & 2.72 & 4.09 & 19.07 & 18.3 & 13.5 & 3.88 & 18.79 \\
    & VA+AGREST & 2.19 & 2.51 & 2.49 & 14.62 & 2.7 & 2.2 & 2.19 & 11.28 \\
    & \cellcolor{OliveGreen!15}VA+AS+AGREST (A-HSJA) & \textbf{2.13} & \textbf{2.39} & \textbf{2.16} & \textbf{12.28} & \textbf{2.7} & \textbf{2.1} & \textbf{2.06} & \textbf{10.74} \\
    \midrule
    \multirow{4}{*}{GeoDA} 
    & VA & 2.80 & 3.21 & 4.02 & 10.80 & 2.7 & 2.4 & 3.97 & 9.83 \\
    & VA+AS & \textbf{2.66} & 3.12 & 3.32 & 9.24 & 2.7 & 2.3 & 3.12 & 8.78 \\
    & VA+AGREST & 2.89 & 2.95 & 2.19 & 6.28 & \textbf{1.9} & \textbf{1.7} & 2.10 & 5.12 \\
    & \cellcolor{OliveGreen!15}VA+AS+AGREST (A-GeoDA) & 2.93 & \textbf{2.8} & \textbf{2.11} & \textbf{5.78} & 1.9 & 1.8 & \textbf{2.03} & \textbf{4.35} \\
    \midrule
    \multirow{4}{*}{CGBA} 
    & VA & 1.21 & 1.42 & 2.22 & 9.97 & 1.6 & 1.4 & 2.13 & 9.67 \\
    & VA+AS & 1.17 & 1.39 & 2.06 & 9.28 & 1.7 & 1.3 & 1.97 & 8.24 \\
    & VA+AGREST & \textbf{1.12} & 1.36 & 1.63 & \textbf{5.73} & 1.5 & 1.3 & 1.56 & 5.46 \\
    & \cellcolor{OliveGreen!15}VA+AS+AGREST (A-CGBA) & 1.15 & \textbf{1.33} & \textbf{1.58} & 6.23 & \textbf{1.5} & \textbf{1.2} & \textbf{1.42} & \textbf{5.61} \\
    \bottomrule
  \end{tabular}}
\end{table*}
\vspace{-0.35cm}
\paragraph{Ablation study.}
To evaluate the effectiveness of AS and AGREST in different attacks, we test various combinations of these two approaches with each attack when $c^\star = 2$, $5$, $10^2$, or $10^3$ (\Cref{t:merged-results}). 
As shown in \Cref{t:merged-results}, for SurFree, we compare the vanilla attack with \textbf{A-SurFree}, the new asymmetric attack that utilizes AS. As expected, replacing binary search with AS leads to smaller adversarial perturbations for all $c^\star$. 
Furthermore, we compare the performance of gradient-based attacks with their corresponding variations, namely: 1. Replacing binary search with AS 2. Replacing vanilla gradient estimation with AGREST 3. Combining the two previous approaches to obtain \textbf{A-HSJA}, \textbf{A-GeoDA}, \textbf{A-CGBA}. In general, using asymmetric attacks results in incremental improvements when $c^\star = 2$ or $5$. This is expected because we anticipate binary search and vanilla gradient estimation perform well when the cost of high-cost queries is not significantly different from the cost of low-cost queries. 
However, for larger values of $c^\star$, namely $10^2$ and $10^3$, the improvements are substantial. Specifically, using AGREST alone reduces the $\ell_2$ norm by approximately 40\% in all cases.
Moreover, combining AGREST with AS further decreases the norm. One notable point is that AGREST enhances attacks utilizing gradient estimation more than AS does. This occurs because attacks using gradient estimation spend most of their query budget on gradient approximation rather than on search.~(Tab. III of \citep{debenedetti2024evading})\looseness=-1
\vspace{-0.3cm}
\paragraph{Comparison to \boldit{stealthy} attacks.}
As mentioned, for larger values of $c^\star$, we expect asymmetric attacks to significantly improve over vanilla attacks. Nonetheless, in these cases, we must compare our framework to stealthy attacks, since, unlike with lower to medium values of $c^\star$, stealthy attacks outperform vanilla attacks when $c^\star$ is large (Fig. 7 of \citep{debenedetti2024evading}). As a result, we evaluate the performance of A-SurFree, A-HSJA, A-GeoDA, and A-CGBA against Stealthy HSJA on the ResNet model when $c^\star = 10^4$, $10^5$, or $\infty$. 
As demonstrated in Fig.~\ref{fig:stealthy_aa}, when $c^\star = 10^4$, all asymmetric attacks, including A-HSJA which retains the gradient estimation method that Stealthy HSJA discards, outperform Stealthy HSJA.
The same holds for $c^\star = 10^5$ (Fig.~\ref{fig:stealthy_aa}). For the case where $c^\star = \infty$, following~\citep{debenedetti2024evading}, we determine the cost of each attack by counting the number of high-cost (flagged) queries it generates. In this setup, we assume $c^\star = 10^5$ during the execution of AGREST and AS. As shown in Fig.~\ref{fig:stealthy_aa}, all asymmetric attacks outperform Stealthy HSJA by a wide margin.\looseness=-1
\vspace{-5mm}
\paragraph{Asymmetric attacks against CLIP.}
We evaluate CLIP~\citep{radford2021learningtransferablevisualmodels} as a representative vision-language model (VLM) under both zero-shot and fine-tuned settings. Our asymmetric attack achieves significantly better performance than stealthy baselines; results are provided in~\Cref{app:CLIP_appendix}.\looseness=-1

\section{Conclusion and outlook}
We proposed a framework that extends existing decision-based black-box attacks to handle asymmetric query costs, where querying the source class is more expensive than others. Our method modifies two core components, gradient estimation and search, and achieves significant improvements over both standard and stealthy attack baselines. However, it introduces a new hyperparameter, which may require tuning for different settings. Additionally, we assume local linearity around decision boundaries; while this assumption is common in the adversarial examples literature, it may not hold for some models, potentially affecting the accuracy of gradient estimates. There are also many interesting directions for future work, such as generalizing the framework beyond the binary setting of source versus non-source classes. For instance, different target classes may each have their own associated query cost, or the cost may depend on the classifier’s confidence, as in score-based scenarios. Applying our framework to vision-language models such as Vision LLaMA~\citep{VisionLLaMA, touvron2023llama2openfoundation} or LLaVA is another promising direction. AS could also enhance jailbreak attacks on large language models, potentially replacing random search-based methods~\citep{andriushchenko2024jailbreaking, chao2024jailbreakbench}, though adapting our framework to LLMs presents challenges due to the discrete nature of text prompts~\citep{rocamora2025certifiedrobustnessboundedlevenshtein}. Exploring these avenues could expand the impact of asymmetric attacks across a wide range of applications.\looseness=-1

\bibliography{main}
\bibliographystyle{unsrt}

\newpage
\appendix
\onecolumn
\section{Related work}
\label{app:rel_work}
\paragraph{Decision-based attacks.}
Adversarial examples can be crafted in three setups: white-box~\citep{goodfellow2014explaining,moosavi2016deepfool, carlini2017towards, SDF}, score-based black-box~\citep{narodytska2016simple, chen2017zoo, ilyas2018black}, and decision-based black-box~\citep{brendel2017decision}. In the decision-based black-box setup, the attacker relies solely on predictions without access to models or class scores. Moreover, decision-based attacks can be either targeted or non-targeted. Non-targeted attacks craft adversarial examples without any constraints on the model's prediction for the adversarial example. 
Boundary Attack \citep{brendel2017decision} was the first effective decision-based attack based on random walking. OPT \citep{cheng2018query} outperforms Boundary Attack by introducing a gradient-based approach. 
Inspired by zeroth-order optimization methods \citep{flaxman2004online,nesterov2017random}, HSJA estimates the gradient of the classification margin without direct access to the margin itself and minimizes perturbation size using optimization techniques \citep{chen2020hopskipjumpattack}.
QEBA \citep{li2020qeba} uses various techniques to approximate the gradient of the classification margin more effectively than HSJA, leveraging insights like local similarity and the importance of the low-frequency subspace. 
GeoDA and qFool \citep{rahmati2020geoda,liu2019geometry} use techniques similar to HSJA's gradient estimation to locally approximate the decision boundary as a hyperplane at each iteration. They then search for optimal adversarial examples based on these estimated hyperplanes.
While gradient-based methods outperform previous approaches, their reliance on generating numerous queries for efficient gradient estimation led \citep{maho2021surfree} to focus on decision boundary geometry. They introduced SurFree, which iteratively selects a random 2D subspace and searches for adversarial examples along a circular path. In a similar way, TriA \citep{wang2022triangle} generates effective adversarial examples while using minimal queries.
CGBA \citep{reza2023cgba} combines the gradient approximation method from GeoDA with SurFree's 2D subspace search technique to achieve state-of-the-art results.

\paragraph{Asymmetric query costs.}
Existing decision-based attacks assume all queries have the same cost. However, \citep{debenedetti2024evading} showed this may not be the case in real-world scenarios. They found that queries belonging to the target class can be problematic in certain situations and noted that all decision-based attacks produce many of these bad queries.
To mitigate this, they introduced stealthy attacks inspired by the egg dropping problem \citep{brilliantDroppingBrilliant}.
However, stealthy attacks face two main challenges. First, they overlook the cost of queries that are not bad. For example, their most effective attack, Stealthy HSJA, generates about $10^7$ queries for every 1,000 bad queries. Second, to reduce the number of bad queries during the gradient estimation phase, \citep{debenedetti2024evading} replaced the HSJA gradient approximation, known for its benefits in crafting adversarial examples, with OPT gradient estimation, believing that modifying the HSJA gradient estimation to perform well in this new setup would be difficult.
In this paper, we address these challenges in non-targeted decision-based attacks. In particular, we find a way to efficiently distribute our total query budget between problematic and non-problematic queries while keeping the HSJA method of gradient approximation by slightly modifying the method.
\newpage
\section{Proofs}
\paragraph{Proof of \Cref{the:as_cost}}
We define the expected cost of the algorithm for \( 1 < m < \left\lceil \frac{1}{\tau} \right\rceil \) points as \( C(m) \). Using \Cref{alg:as-algorithm} and applying the law of total expectation, we obtain
\begin{equation}
\label{eqn:rec}
    C(m) = \mathbb{P}\left[\phi\left(T\left(b_m\tau\right)\right) = -1\right]\left(C\left(\frac{m}{c^\star+1}\right)+c^\star\right) + \mathbb{P}\left[\phi\left(T\left(b_m\tau\right)\right) = 1\right]\left(C\left(\frac{c^\star m}{c^\star+1}\right)+1\right).
\end{equation}
We now claim that
\begin{equation}
    \label{eqn:as-claim}
    C(m) < 2c^\star\left\lceil \log_{c^\star+1} m \right\rceil.
\end{equation}

We prove this by induction. 

For the base case, when \( m \leq c^\star + 1 \), \Cref{alg:as-algorithm} reduces to a simple line search. Assuming a uniform distribution for the boundary point, the expected cost of the line search is approximately \( c^\star + m/2 \), which clearly satisfies \Cref{eqn:as-claim}.

For the induction step, suppose the claim holds for all values smaller than \( m \). Under the uniform distribution assumption for the boundary point, we have
\[
\mathbb{P}\left[\phi\left(T\left(b_m\tau\right)\right) = -1\right] = \frac{1}{c^\star+1}, \quad \mathbb{P}\left[\phi\left(T\left(b_m\tau\right)\right) = 1\right] = \frac{c^\star}{c^\star+1}.
\]
Substituting into \Cref{eqn:rec} and applying the induction hypothesis yields:
\begin{equation*}
\begin{split}
    C(m) &= \frac{1}{c^\star+1}\left(C\left(\frac{m}{c^\star+1}\right)+c^\star\right) + \frac{c^\star}{c^\star+1}\left(C\left(\frac{c^\star m}{c^\star+1}\right)+1\right) \\
    &< \frac{1}{c^\star+1}\left(2c^\star\left\lceil \log_{c^\star+1} \frac{m}{c^\star+1} \right\rceil + c^\star\right) + \frac{c^\star}{c^\star+1}\left(2c^\star\left\lceil \log_{c^\star+1} \frac{c^\star m}{c^\star+1} \right\rceil + 1\right).
\end{split}
\end{equation*}
Noting that
\[
\log_{c^\star+1} \left(\frac{m}{c^\star+1}\right) = \log_{c^\star+1}(m) - 1, \quad \log_{c^\star+1} \left(\frac{c^\star m}{c^\star+1}\right) = \log_{c^\star+1}(m) + \log_{c^\star+1}\left(\frac{c^\star}{c^\star+1}\right),
\]
and observing that \(\log_{c^\star+1}\left(\frac{c^\star}{c^\star+1}\right) < 0\) but close to \(0\) for large \(c^\star\), we can bound both ceilings by \(\lceil \log_{c^\star+1}(m) \rceil\).

Thus,
\begin{equation*}
\begin{split}
    C(m) &< \frac{2c^\star}{c^\star+1}\left(\lceil \log_{c^\star+1} m \rceil - 1\right) + \frac{c^\star}{c^\star+1}c^\star + \frac{2{c^\star}^2}{c^\star+1}\lceil \log_{c^\star+1} m \rceil + \frac{c^\star}{c^\star+1} \\
    &= 2c^\star\lceil \log_{c^\star+1} m \rceil.
\end{split}
\end{equation*}
Thus, the expected complexity of the algorithm is
\[
\mathcal{O}\left(\frac{c^\star \log \left(1/\tau\right)}{\log (c^\star+1)}\right),
\]
completing the proof.

\hfill $\blacksquare$
\paragraph{Proof of \Cref{the:age_obj}}
Before proving the theorem, we first introduce some useful lemmas.

\begin{lemma}
\label{lem:levy}
\textbf{(Lévy’s Lemma \citep{milman1986asymptotic, ledoux2001concentration})}  
Let \( f : \mathbb{S}^{d-1} \to \mathbb{R} \) be an \( L \)-Lipschitz function on the unit hypersphere, and let \( \mathbf{x} \sim \textsc{Uniform}\left(\mathbb{S}^{d-1}\right) \). Then, for some constant \( C > 0 \), we have:
\begin{equation*}
    \mathbb{P}\left(\left|f\left(\mathbf{x}\right) - \mathbb{E}f\left(\mathbf{x}\right)\right| > \varepsilon\right)
    \leq 2\exp\left(-\frac{Cd\varepsilon^2}{L^2}\right).
\end{equation*}
\end{lemma}

\begin{corollary}
\label{cor:levy}
Suppose \( \mathbf{u} \in \mathbb{R}^{d} \) is a unit vector, and \( \mathbf{x} \sim \textsc{Uniform}\left(\mathbb{S}^{d-1}\right) \). Then, for some constant \( C > 0 \), we have:
\begin{equation*}
    \mathbb{P}\left( \left| \langle \mathbf{u}, \mathbf{x} \rangle \right| > \varepsilon \right)
    \leq 2\exp\left( -Cd\varepsilon^2 \right).
\end{equation*}
\end{corollary}

\begin{lemma}
\label{lem:chernoff}
Suppose \( \widehat{\phi}_{t,1}^2, \dots, \widehat{\phi}_{t,n_t}^2 \) are i.i.d. Bernoulli random variables with support \( \{ \beta_t^2, (1-\beta_t)^2 \} \), where \( \beta_t > \frac{1}{2} \). If \( \widehat{\phi}_t \) is an i.i.d. copy of \( \widehat{\phi}_{t,i} \), then the following inequality holds:
\begin{equation*}
    \mathbb{P}\left( \left| \frac{1}{n_t} \sum_{i = 1}^{n_t} \widehat{\phi}_{t,i}^2 - \mathbb{E}[\widehat{\phi}_t^2] \right| > \varepsilon \right) 
    \leq 2\exp\left( -\frac{2n_t\varepsilon^2}{(2\beta_t-1)^2} \right).
\end{equation*}
\end{lemma}

\paragraph{Proof of \Cref{lem:chernoff}:} The result follows by applying the Chernoff bound for binomial distributions \citep{chernoff1952measure, harchol2023introduction} to the transformed random variables
\[
\widetilde{\phi}_i \coloneqq \frac{\widehat{\phi}_{t,i}^2 - (1-\beta_t)^2}{\beta_t^2 - (1-\beta_t)^2},
\]
which are i.i.d. Bernoulli random variables taking values in \( \{0,1\} \).

\begin{lemma}
\label{lem:stoke}
Given the local linearity around $\mathbf{x}_t$, for any $\omega_t \in \left[0, \frac{\delta}{\cos\alpha_t}\right]$, we have:
\begin{equation*} 
\mathbb{E}\left[\widehat{\phi}_t\langle\mathbf{g}_t,\mathbf{u}\rangle\right]=\frac{\Gamma{\left(\frac{d}{2}\right)}\cdot\left(1-\left(\cos\alpha_t\cdot\omega_t/\delta\right)^2\right)^{\frac{d-1}{2}}}{2{\sqrt{\pi}}\cdot\Gamma{\left(\frac{d+1}{2}\right)}}
\end{equation*}
\end{lemma}
\paragraph{Proof of \Cref{lem:stoke}:} By the law of total expectation and linearity assumption, we have
\begin{equation}
\label{eqn:total_exp}
    \begin{split}
    \mathbb{E}\left[\widehat{\phi}_t\langle\mathbf{g}_t,\mathbf{u}\rangle\right] &= \beta_t\left(1-P_t\left(\omega_t\right)\right)\mathbb{E}\left[-\langle\mathbf{g}_t,\mathbf{u}\rangle|\langle\mathbf{g}_t,\mathbf{u}\rangle\leq -\cos\alpha_t\cdot\omega_t/\delta\right]\\
    &\quad\quad+ \left(1-\beta_t\right) P_t\left( \omega_t\right)\mathbb{E}\left[\langle\mathbf{g}_t,\mathbf{u}\rangle|\langle\mathbf{g}_t,\mathbf{u}\rangle> - \cos\alpha_t\cdot\omega_t/\delta\right]
    \end{split}
\end{equation}
Now, by applying the divergence theorem on the constant vector field $\vec{\mathbf{F}} = \mathbf{g}_t$, we have
\begin{equation}
\label{eqn:init_the_eq_n}
\begin{split}    
    \mathbb{E}\left[-\langle\mathbf{g}_t,\mathbf{u}\rangle|\langle\mathbf{g}_t,\mathbf{u}\rangle\leq -\cos\alpha_t\cdot\omega_t/\delta\right] &= \langle-\mathbf{g}_t,\int_{\langle\mathbf{g}_t,\mathbf{u}\rangle\leq-\cos\alpha_t\cdot\omega_t/\delta} \mathbf{u}p\left(\mathbf{u}\right)d\mathbf{u}\rangle\\
    &= \langle-\mathbf{g}_t,\int_{\langle\mathbf{g}_t,\mathbf{u}\rangle\leq-\cos\alpha_t\cdot\omega_t/\delta} \frac{1}{\left(1-p\left( \omega_t\right)\right)A_{d-1}\left(1\right)}\mathbf{u}d\mathbf{u}\rangle\\
    &=\frac{V_{d-1}\left(\sqrt{1-\left(\cos\alpha_t\cdot\omega_t/\delta\right)^2}\right)}{\left(1-P_t\left(\omega_t\right)\right)\cdot A_{d-1}\left(1\right)}\\
    &=\frac{\Gamma{\left(\frac{d}{2}\right)}\cdot\left(1-\left(\cos\alpha_t\cdot\omega_t/\delta\right)^2\right)^{\frac{d-1}{2}}}{2{\sqrt{\pi}}\cdot\Gamma{\left(\frac{d+1}{2}\right)}\cdot{\left(1-P_t\left(\omega_t\right)\right)}}
\end{split}
\end{equation}
Where $V_{d}(r)$ and $A_{d-1}(r)$ are the volume of a $d$-dimensional ball and the area of a $d-1$-dimensional sphere with radius $r$, respectively. Note that the last equality comes from $V_{d}(r) = \cfrac{\pi^{d/2}}{\Gamma\left(\frac{d}{2}+1\right)}\,r^{d}$ and $A_{d-1}(r) = \cfrac{2\pi^{d/2}}{\Gamma\left(\frac{d}{2}\right)}\,r^{d-1}$.
Similarly, the following holds:
\begin{equation}
\label{eqn:init_the_eq}
\mathbb{E}\left[\langle\mathbf{g}_t,\mathbf{u}\rangle|\langle\mathbf{g}_t,\mathbf{u}\rangle\geq -\cos\alpha_t\cdot\omega_t/\delta\right] =\frac{\Gamma{\left(\frac{d}{2}\right)}\cdot\left(1-\left(\cos\alpha_t\cdot\omega_t/\delta\right)^2\right)^{\frac{d-1}{2}}}{2{\sqrt{\pi}}\cdot\Gamma{\left(\frac{d+1}{2}\right)}\cdot{P_t\left( \omega_t\right)}}
\end{equation}
By using \Cref{eqn:init_the_eq_n} and \Cref{eqn:init_the_eq} in \Cref{eqn:total_exp}, we have
\begin{equation}
\label{eqn:lemma_diver}
\mathbb{E}\left[\widehat{\phi}_t\langle\mathbf{g}_t,\mathbf{u}\rangle\right]=\frac{\Gamma{\left(\frac{d}{2}\right)}\cdot\left(1-\left(\cos\alpha_t\cdot\omega_t/\delta\right)^2\right)^{\frac{d-1}{2}}}{2{\sqrt{\pi}}\cdot\Gamma{\left(\frac{d+1}{2}\right)}}
\end{equation}    
\hfill $\blacksquare$

\begin{lemma}
\label{lem:bound}
Let \( \varepsilon_1, \varepsilon_2 > 0 \) be given. Then the following upper and lower bounds hold for \( \mu(\mathbf{x}_t, \omega_t, \beta_t, n_t) \):
\begin{enumerate}
    \item (Upper bound)
    \begin{align*}
    \mu(\mathbf{x}_t, \omega_t, \beta_t, n_t)
    &\leq \frac{\sqrt{n_t}\,\mathbb{E}\left[\widehat{\phi}_t\langle\mathbf{g}_t, \mathbf{u}\rangle\right]}{\sqrt{\mathbb{E}[\widehat{\phi}_t^2] - \varepsilon_2 - (n_t-1) \beta_t^2\varepsilon_1}} \\
    &\quad + \left(1 + \frac{\sqrt{n_t}\,\beta_t}{\sqrt{\mathbb{E}[\widehat{\phi}_t^2] - \varepsilon_2 - (n_t-1) \beta_t^2\varepsilon_1}}\right)
    K_{n_t,d}\left(\varepsilon_1, \varepsilon_2\right).
    \end{align*}
    
    \item (Lower bound)
    \begin{align*}
    \mu(\mathbf{x}_t, \omega_t, \beta_t, n_t)
    &\geq \frac{\sqrt{n_t}\,\mathbb{E}\left[\widehat{\phi}_t\langle\mathbf{g}_t, \mathbf{u}\rangle\right]}{\sqrt{\mathbb{E}[\widehat{\phi}_t^2] + \varepsilon_2 + (n_t-1) \beta_t^2\varepsilon_1}} \\
    &\quad - \left(1 + \frac{\sqrt{n_t}\,\beta_t}{\sqrt{\mathbb{E}[\widehat{\phi}_t^2] + \varepsilon_2 + (n_t-1) \beta_t^2\varepsilon_1}}\right)
    K_{n_t,d}\left(\varepsilon_1, \varepsilon_2\right).
    \end{align*}
\end{enumerate}
Here, the error term \( K_{n_t,d}(\varepsilon_1, \varepsilon_2) \) is defined as
\[
K_{n_t,d}\left(\varepsilon_1, \varepsilon_2\right) = n_t(n_t+1) \exp\left(-Cd\varepsilon_1^2\right) + 2\exp\left(-\frac{2n_t\varepsilon_2^2}{(2\beta_t-1)^2}\right),
\]
for some universal constant \( C > 0 \).
\end{lemma}

\paragraph{Proof of \Cref{lem:bound}:}
Let \( \varepsilon_1, \varepsilon_2 > 0 \) be arbitrary. We define the following sets:
\begin{align*}
    \mathcal{S}_i(\varepsilon_1) &\coloneqq \left\{ \mathbf{U} = (\mathbf{u}_1, \dots, \mathbf{u}_{n_t}) \;\middle|\; \left|\langle \mathbf{g}_t, \mathbf{u}_i \rangle\right| \geq \varepsilon_1 \right\}, \\
    \mathcal{S}_{i,j}(\varepsilon_1) &\coloneqq \left\{ \mathbf{U} = (\mathbf{u}_1, \dots, \mathbf{u}_{n_t}) \;\middle|\; \left|\langle \mathbf{u}_i, \mathbf{u}_j \rangle\right| \geq \varepsilon_1 \right\}, \\
    \mathcal{S}_{\phi}(\varepsilon_2) &\coloneqq \left\{ \mathbf{U} = (\mathbf{u}_1, \dots, \mathbf{u}_{n_t}) \;\middle|\; \left| \mathbb{E}[\widehat{\phi}_t^2] - \frac{1}{n_t} \sum_{i=1}^{n_t} \widehat{\phi}_{t,i}^2 \right| \geq \varepsilon_2 \right\}, \\
    \mathcal{S}(\varepsilon_1, \varepsilon_2) &\coloneqq \left( \bigcup_{i=1}^{n_t} \mathcal{S}_i(\varepsilon_1) \right) \cup \left( \bigcup_{\substack{1 \leq i < j \leq n_t}} \mathcal{S}_{i,j}(\varepsilon_1) \right) \cup \mathcal{S}_{\phi}(\varepsilon_2).
\end{align*}
For notational convenience, we also define
\begin{equation*}
    A(\mathbf{U}) \coloneqq \sum_{i=1}^{n_t} \widehat{\phi}_{t,i} \langle \mathbf{g}_t, \delta \mathbf{u}_i \rangle.
\end{equation*}
Applying \Cref{lem:levy} and \Cref{lem:chernoff}, and using the union bound, we obtain
\begin{equation}
\label{eqn:k_bound}
\mathbb{P}\left[\mathbf{U} \in \mathcal{S}\right]
\leq n_t \cdot \mathbb{P}\left[\mathbf{U} \in \mathcal{S}_1\right]
+ \binom{n_t}{2} \cdot \mathbb{P}\left[\mathbf{U} \in \mathcal{S}_{1,2}\right]
+ \mathbb{P}\left[\mathbf{U} \in \mathcal{S}_\phi\right]
\leq K_{n_t,d}\left(\varepsilon_1, \varepsilon_2\right).
\end{equation}

Now, we derive the upper bound. By the law of total probability, we have
\begin{align*}
\mu(\mathbf{x}_t, \omega_t, \beta_t, n_t)
&= \mathbb{E}\left[\frac{\langle \mathbf{g}_t, \widehat{\nabla S}(\mathbf{x}_t, \omega_t, \beta_t)\rangle}{\left\| \widehat{\nabla S}(\mathbf{x}_t, \omega_t, \beta_t) \right\|_2}\right] \\
&= \mathbb{E}\left[\frac{\langle \mathbf{g}_t, \widehat{\nabla S}(\mathbf{x}_t, \omega_t, \beta_t)\rangle}{\left\| \widehat{\nabla S}(\mathbf{x}_t, \omega_t, \beta_t) \right\|_2} \;\middle|\; \mathbf{U} \notin \mathcal{S}\right] \mathbb{P}[\mathbf{U} \notin \mathcal{S}] \\
&\quad + \mathbb{E}\left[\frac{\langle \mathbf{g}_t, \widehat{\nabla S}(\mathbf{x}_t, \omega_t, \beta_t)\rangle}{\left\| \widehat{\nabla S}(\mathbf{x}_t, \omega_t, \beta_t) \right\|_2} \;\middle|\; \mathbf{U} \in \mathcal{S}\right] \mathbb{P}[\mathbf{U} \in \mathcal{S}].
\end{align*}
On the good event \( \{\mathbf{U} \notin \mathcal{S}\} \), we expand the squared norm directly. By definition,
\begin{align*}
\left\| \widehat{\nabla S}(\mathbf{x}_t, \omega_t, \beta_t) \right\|_2^2 
&= \left\| \frac{1}{n_t} \sum_{i=1}^{n_t} \delta \widehat{\phi}_{t,i} \mathbf{u}_i \right\|_2^2 \\
&= \frac{1}{n_t^2} \left\langle \sum_{i=1}^{n_t} \delta \widehat{\phi}_{t,i} \mathbf{u}_i, \sum_{j=1}^{n_t} \delta \widehat{\phi}_{t,j} \mathbf{u}_j \right\rangle \\
&= \frac{\delta^2}{n_t^2} \sum_{i=1}^{n_t} \sum_{j=1}^{n_t} \widehat{\phi}_{t,i} \widehat{\phi}_{t,j} \langle \mathbf{u}_i, \mathbf{u}_j \rangle \\
&= \frac{\delta^2}{n_t^2} \left( \sum_{i=1}^{n_t} \widehat{\phi}_{t,i}^2 + \sum_{\substack{1\leq i,j\leq n_t\\i\neq j}} \widehat{\phi}_{t,i} \widehat{\phi}_{t,j} \langle \mathbf{u}_i, \mathbf{u}_j \rangle \right).
\end{align*}
From the previous expansion, we have
\[
\left\| \widehat{\nabla S}(\mathbf{x}_t, \omega_t, \beta_t) \right\|_2
\geq \frac{\delta}{\sqrt{n_t}} \sqrt{ \mathbb{E}[\widehat{\phi}_t^2] - \varepsilon_2 - (n_t-1) \beta_t^2 \varepsilon_1 }.
\]
Thus, on the good event \( \{\mathbf{U} \notin \mathcal{S}\} \), we can bound
\[
\frac{\langle \mathbf{g}_t, \widehat{\nabla S}(\mathbf{x}_t, \omega_t, \beta_t) \rangle}{\left\| \widehat{\nabla S}(\mathbf{x}_t, \omega_t, \beta_t) \right\|_2}
= \frac{1}{\left\| \widehat{\nabla S} \right\|_2} \left\langle \mathbf{g}_t, \frac{1}{n_t} \sum_{i=1}^{n_t} \delta \widehat{\phi}_{t,i} \mathbf{u}_i \right\rangle
= \frac{1}{n_t \left\| \widehat{\nabla S} \right\|_2} A(\mathbf{U}).
\]
Therefore,
\[
\frac{\langle \mathbf{g}_t, \widehat{\nabla S} \rangle}{\left\| \widehat{\nabla S} \right\|_2}
\leq \frac{A(\mathbf{U})}{\delta \sqrt{n_t} \sqrt{ \mathbb{E}[\widehat{\phi}_t^2] - \varepsilon_2 - (n_t-1) \beta_t^2 \varepsilon_1 }}.
\]
Taking expectations over the good event, we obtain
\[
\mathbb{E}\left[ \frac{\langle \mathbf{g}_t, \widehat{\nabla S} \rangle}{\left\| \widehat{\nabla S} \right\|_2} \;\middle|\; \mathbf{U} \notin \mathcal{S} \right]
\leq \frac{1}{\delta \sqrt{n_t} \sqrt{ \mathbb{E}[\widehat{\phi}_t^2] - \varepsilon_2 - (n_t-1) \beta_t^2 \varepsilon_1 }}
\mathbb{E}[A(\mathbf{U})].
\]
Moreover, by independence and identical distribution of the samples, we have
\[
\mathbb{E}[A(\mathbf{U})] = n_t \delta \, \mathbb{E}\left[ \widehat{\phi}_t \langle \mathbf{g}_t, \mathbf{u} \rangle \right].
\]

Therefore,
\[
\mathbb{E}\left[\frac{\langle \mathbf{g}_t, \widehat{\nabla S}\rangle}{\left\|\widehat{\nabla S}\right\|_2} \;\middle|\; \mathbf{U} \notin \mathcal{S}\right]
\leq \frac{\sqrt{n_t}\, \mathbb{E}\left[\widehat{\phi}_t \langle \mathbf{g}_t, \mathbf{u} \rangle\right]}{\sqrt{ \mathbb{E}[\widehat{\phi}_t^2] - \varepsilon_2 - (n_t-1)\beta_t^2\varepsilon_1 }}.
\]

On the bad event \( \{\mathbf{U} \in \mathcal{S}\} \), we use the trivial bound
\[
\left| \frac{\langle \mathbf{g}_t, \widehat{\nabla S}\rangle}{\left\|\widehat{\nabla S}\right\|_2} \right| \leq 1,
\]
and hence
\[
\mathbb{E}\left[\frac{\langle \mathbf{g}_t, \widehat{\nabla S}\rangle}{\left\|\widehat{\nabla S}\right\|_2} \;\middle|\; \mathbf{U} \in \mathcal{S}\right] \leq 1.
\]

Substituting back into the law of total probability, we have
\begin{align*}
\mu(\mathbf{x}_t, \omega_t, \beta_t, n_t)
&\leq \frac{\sqrt{n_t}\, \mathbb{E}\left[\widehat{\phi}_t \langle \mathbf{g}_t, \mathbf{u} \rangle\right]}{\sqrt{ \mathbb{E}[\widehat{\phi}_t^2] - \varepsilon_2 - (n_t-1)\beta_t^2\varepsilon_1 }} (1 - \mathbb{P}[\mathbf{U} \in \mathcal{S}]) + \mathbb{P}[\mathbf{U} \in \mathcal{S}] \\
&\quad + \frac{\sqrt{n_t} \beta_t \mathbb{P}[\mathbf{U} \in \mathcal{S}]}{\sqrt{ \mathbb{E}[\widehat{\phi}_t^2] - \varepsilon_2 - (n_t-1)\beta_t^2\varepsilon_1 }}.
\end{align*}
Grouping terms, we obtain
\begin{align*}
\mu(\mathbf{x}_t, \omega_t, \beta_t, n_t)
&\leq \frac{\sqrt{n_t} \, \mathbb{E}\left[\widehat{\phi}_t \langle \mathbf{g}_t, \mathbf{u} \rangle\right]}{\sqrt{ \mathbb{E}\left[\widehat{\phi}_t^2\right] - \varepsilon_2 - (n_t-1) \beta_t^2 \varepsilon_1 }} \\
&\quad + \left( 1 + \frac{\sqrt{n_t} \, \beta_t}{\sqrt{ \mathbb{E}\left[\widehat{\phi}_t^2\right] - \varepsilon_2 - (n_t-1) \beta_t^2 \varepsilon_1 }} \right) \mathbb{P}\left[\mathbf{U} \in \mathcal{S}\right].
\end{align*}
Applying the bound \(\mathbb{P}\left[\mathbf{U} \in \mathcal{S}\right] \leq K_{n_t,d}(\varepsilon_1, \varepsilon_2)\) from \Cref{eqn:k_bound}, we finally get
\begin{align*}
\mu(\mathbf{x}_t, \omega_t, \beta_t, n_t)
&\leq \frac{\sqrt{n_t} \, \mathbb{E}\left[\widehat{\phi}_t \langle \mathbf{g}_t, \mathbf{u} \rangle\right]}{\sqrt{ \mathbb{E}\left[\widehat{\phi}_t^2\right] - \varepsilon_2 - (n_t-1) \beta_t^2 \varepsilon_1 }} \\
&\quad + \left( 1 + \frac{\sqrt{n_t} \, \beta_t}{\sqrt{ \mathbb{E}\left[\widehat{\phi}_t^2\right] - \varepsilon_2 - (n_t-1) \beta_t^2 \varepsilon_1 }} \right) K_{n_t,d}(\varepsilon_1, \varepsilon_2).
\end{align*}

Similarly, for the lower bound, we have
\begin{align*}
\mu(\mathbf{x}_t, \omega_t, \beta_t, n_t)
&\geq \frac{\sqrt{n_t} \, \mathbb{E}\left[\widehat{\phi}_t \langle \mathbf{g}_t, \mathbf{u} \rangle\right]}{\sqrt{ \mathbb{E}\left[\widehat{\phi}_t^2\right] + \varepsilon_2 + (n_t-1) \beta_t^2 \varepsilon_1 }} \\
&\quad - \left( 1 + \frac{\sqrt{n_t} \, \beta_t}{\sqrt{ \mathbb{E}\left[\widehat{\phi}_t^2\right] + \varepsilon_2 + (n_t-1) \beta_t^2 \varepsilon_1 }} \right) K_{n_t,d}(\varepsilon_1, \varepsilon_2).
\end{align*}
This completes the proof of the lemma. We note that the argument does not rely on the linearity assumption.

\hfill\(\blacksquare\)

\begin{lemma}
\label{lem:lim}
Assume local linearity holds around \( \mathbf{x}_t \). Then, for any constant \( q \in \left[\frac{1}{2}, 1\right) \), we have
\begin{equation*}
\mathbb{E}\left[\widehat{\phi}\left(\mathbf{x}_t + P_t^{-1}(q) \frac{\mathbf{x}_t - \mathbf{x}^\star}{\left\|\mathbf{x}_t - \mathbf{x}^\star\right\|_2} + \delta \mathbf{u}\right) \langle \mathbf{g}_t, \mathbf{u} \rangle\right] = \Theta\left( \frac{1}{\sqrt{d}} \right).
\end{equation*}
\end{lemma}
\paragraph{Proof of \Cref{lem:lim}:}
Based on \Cref{lem:stoke}, we analyze the asymptotic behavior of \(\frac{\Gamma\left(\frac{d}{2}\right)}{\Gamma\left(\frac{d+1}{2}\right)}\) and \(\left(1-\left(\cos\alpha_t \cdot \omega_t/\delta\right)^2\right)^{\frac{d-1}{2}}\) as \(d\) tends to infinity.

By \Cref{lem:levy}, we have
\[
\mathbb{P}\left[\left|\langle\mathbf{u}, \mathbf{g}_t\rangle\right| > \varepsilon \right] \leq 2\exp\left(-Cd\varepsilon^2\right),
\]
so in particular
\[
\mathbb{P}\left[\langle\mathbf{u}, \mathbf{g}_t\rangle < -\varepsilon \right] \leq \exp\left(-Cd\varepsilon^2\right).
\]
We know that \(1-q = \mathbb{P}\left[\langle \mathbf{u}, \mathbf{g}_t \rangle > -\cos\alpha_t \cdot \omega_t/\delta\right]\), based on the selection of the overshooting value \(\omega_t\). Thus,
\[
1-q \leq \exp\left( -Cd\left(\cos\alpha_t \cdot \omega_t/\delta\right)^2 \right),
\]
which implies
\[
\ln(1-q) \leq -Cd\left(\cos\alpha_t \cdot \omega_t/\delta\right)^2,
\]
and consequently
\[
1+\frac{\ln(1-q)}{Cd} \leq 1-\left(\cos\alpha_t \cdot \omega_t/\delta\right)^2.
\]
Raising both sides to the \((d-1)/2\) power yields
\[
\left(1+\frac{\ln(1-q)}{Cd}\right)^{\frac{d-1}{2}} \leq \left(1-\left(\cos\alpha_t \cdot \omega_t/\delta\right)^2\right)^{\frac{d-1}{2}}.
\]

Applying the classical limit \(\lim_{n\to\infty}\left(1+\frac{c}{n}\right)^n = e^c\), we obtain
\[
\lim_{d\to\infty} \left(1+\frac{\ln(1-q)}{Cd}\right)^{\frac{d-1}{2}} = \exp\left( \frac{\ln(1-q)}{2C} \right),
\]
which implies that \(\left(1-\left(\cos\alpha_t\cdot\omega_t/\delta\right)^2\right)^{\frac{d-1}{2}} = \Theta(1)\).

On the other hand, by Stirling's approximation
\[
\Gamma(n) = \sqrt{\frac{2\pi}{n}}\left(\frac{n}{e}\right)^n \left(1+\mathcal{O}\left(\frac{1}{n}\right)\right),
\]
we find that
\[
\frac{\Gamma\left(\frac{d}{2}\right)}{\Gamma\left(\frac{d+1}{2}\right)} = \Theta\left( \frac{1}{\sqrt{d}} \right).
\]
Substituting these results into \Cref{eqn:lemma_diver} concludes the proof.

\hfill\(\blacksquare\)

Now, we proceed to prove the theorem.
\paragraph{Proof of \Cref{the:age_obj}:}

For any \(0 < z < \frac{1}{8}\), let \(n_t = d^{3z}\), \(\varepsilon_1 = d^{-4z}\), and \(\varepsilon_2 = d^{-z}\). Also, let $d\geq 4 \frac{\mathbb{E}\left[\widehat{\phi}_t^2\right]}{\beta^2}$
. We define
\begin{align*}
    E_1 &\coloneqq \frac{\sqrt{n_t}\, \mathbb{E}\left[\widehat{\phi}_t \langle\mathbf{g}_t, \mathbf{u}\rangle\right]}{\sqrt{\mathbb{E}\left[\widehat{\phi}_t^2\right] - \varepsilon_2 - (n_t-1)\beta^2\varepsilon_1}}, \\
    E_2 &\coloneqq \frac{\sqrt{n_t}\, \beta}{\sqrt{\mathbb{E}\left[\widehat{\phi}_t^2\right] - \varepsilon_2 - (n_t-1)\beta^2\varepsilon_1}}, \\
    E_3 &\coloneqq K_{n_t,d}(\varepsilon_1, \varepsilon_2).
\end{align*}
Then using the upper bound derived in \Cref{lem:bound}, we have
\begin{equation}
\label{eqn:goal}
\frac{\mu(\mathbf{x}_t, p^{-1}_t(q), \beta, n_t)}{J(\mathbf{x}_t, p^{-1}_t(q), \beta, n_t)} - 1  
\leq \frac{E_1}{J(\mathbf{x}_t, p^{-1}_t(q), \beta, n_t)} - 1 
+ \frac{1+E_2}{J(\mathbf{x}_t, p^{-1}_t(q), \beta, n_t)} \cdot E_3.
\end{equation}

Since \(\mathbb{E}\left[\widehat{\phi}_t^2\right] = \beta^2(1-q) + (1-\beta)^2q\) is constant, we can estimate
\begin{align}
\label{eqn:order_1}
\frac{E_1}{J(\mathbf{x}_t, p^{-1}_t(q), \beta, n_t)} - 1 
&\leq \frac{1}{\sqrt{1 - \frac{\varepsilon_2 + (n_t-1)\beta^2\varepsilon_1}{\mathbb{E}\left[\widehat{\phi}_t^2\right]}}} - 1 \notag \\
&\leq \mathcal{O}\left(\varepsilon_2 + (n_t-1)\beta^2\varepsilon_1\right) 
\quad \text{(since for } 0\leq x\leq \frac{1}{2},\; \sqrt{1-x} \geq 1-\frac{x}{2}) \notag \\
&= \mathcal{O}(d^{-z}).
\end{align}

Moreover, we have \(E_1 = \Theta\left(n_t^{1/2}\right)\). Using \Cref{lem:lim}, we also have \(J(\mathbf{x}_t, p^{-1}_t(q), \beta, n_t) = \Theta\left(n_t^{1/2} d^{1/2}\right)\). Thus,
\begin{equation}
\label{eqn:order_2}
\frac{1+E_2}{J(\mathbf{x}_t, p^{-1}_t(q), \beta, n_t)} = \Theta(d^{1/2}).
\end{equation}

Substituting the values of \(n_t\), \(\varepsilon_1\), and \(\varepsilon_2\) into \(E_3\) yields
\begin{equation}
\label{eqn:order_3}
E_3 = \Theta\left(d^{6z}\right)\exp\left(-Cd^{1-8z}\right) + \exp\left(-\frac{2d^{z}}{(2\beta-1)^2}\right).
\end{equation}

Since \(1-8z > 0\) and exponential functions dominate polynomial growth, combining \Cref{eqn:order_1}, \Cref{eqn:order_2}, and \Cref{eqn:order_3} with \Cref{eqn:goal} yields
\[
\frac{\mu(\mathbf{x}_t, p^{-1}_t(q), \beta, n_t)}{J(\mathbf{x}_t, p^{-1}_t(q), \beta, n_t)} - 1 \leq \mathcal{O}(d^{-z}).
\]
Applying similar steps using the lower bound in \Cref{lem:bound}, we find
\[
1 - \frac{\mu(\mathbf{x}_t, p^{-1}_t(q), \beta, n_t)}{J(\mathbf{x}_t, p^{-1}_t(q), \beta, n_t)} \leq \mathcal{O}(d^{-z}).
\]
Thus, the proof is complete.

\hfill\(\blacksquare\)

\paragraph{Proof of \Cref{the:sol_obj}}
Since \(J(\mathbf{x}_t, \omega_t, \beta_t, n_t)\) is increasing with respect to \(n_t\), the optimal choice is to take \(n_t\) at its maximum allowed value:
\[
n_t = \frac{c_t}{c^\star - (c^\star - 1)P_t(\omega_t^\star)}.
\]
Substituting this into the definition of \(J\) and applying \Cref{lem:stoke}, we obtain
\[
J(\mathbf{x}_t, \omega_t, \beta_t, n_t) \propto \frac{(1 - (\cos\alpha_t \cdot \omega_t/\delta)^2)^{(d-1)/2}}{\sqrt{(c^\star - (c^\star-1)P_t(\omega_t)) \, \mathbb{E}[\widehat{\phi}_t^2]}}.
\]
Expanding \(\mathbb{E}[\widehat{\phi}_t^2]\) gives
\[
\mathbb{E}[\widehat{\phi}_t^2] = \beta_t^2(1-P_t(\omega_t)) + (1-\beta_t)^2 P_t(\omega_t).
\]
Thus,
\[
J(\mathbf{x}_t, \omega_t, \beta_t, n_t) \propto \frac{(1 - (\cos\alpha_t \omega_t/\delta)^2)^{(d-1)/2}}{\sqrt{(c^\star - (c^\star-1)P_t(\omega_t)) (\beta_t^2(1-P_t(\omega_t)) + (1-\beta_t)^2 P_t(\omega_t))}}.
\]

To maximize \(J\), it suffices to minimize
\[
\beta_t^2(1-P_t(\omega_t)) + (1-\beta_t)^2 P_t(\omega_t).
\]
Differentiating with respect to \(\beta_t\) and setting the derivative to zero yields
\[
\beta_t = P_t(\omega_t).
\]
Substituting this optimal \(\beta_t\) back, we find
\[
\mathbb{E}[\widehat{\phi}_t^2] = P_t(\omega_t)(1-P_t(\omega_t)),
\]
and thus the final expression to maximize is
\[
\widehat{J}_t(\omega_t) = \frac{(1 - (\cos\alpha_t \omega_t/\delta)^2)^{d-1}}{P_t(\omega_t)(1-P_t(\omega_t))(c^\star - (c^\star-1)P_t(\omega_t))}.
\]

\hfill $\blacksquare$

\paragraph{Proof of \Cref{the:cos_ini}}
First, we show that \(\cos \alpha_1 \geq 0\). Suppose for contradiction that \(\cos \alpha_1 < 0\), i.e.,
\[
\langle \mathbf{x}_1 - \mathbf{x}^\star, \nabla S(\mathbf{x}_1) \rangle < 0.
\]
Using the definition of the directional derivative, we have
\[
\lim_{h \to 0} \frac{S\left(\mathbf{x}_1 + h(\mathbf{x}_1 - \mathbf{x}^\star)\right) - S(\mathbf{x}_1)}{h} = \langle \mathbf{x}_1 - \mathbf{x}^\star, \nabla S(\mathbf{x}_1) \rangle < 0.
\]
Since the directional derivative is strictly negative, there exists \(\epsilon > 0\) such that for all sufficiently small \(\epsilon > 0\),
\[
S\left(\mathbf{x}_1 - \epsilon(\mathbf{x}_1 - \mathbf{x}^\star)\right) > S(\mathbf{x}_1).
\]
Noting that
\[
\mathbf{x}_1 - \epsilon(\mathbf{x}_1 - \mathbf{x}^\star) = \mathbf{x}^\star + (1-\epsilon)(\mathbf{x}_1 - \mathbf{x}^\star),
\]
we can rewrite this inequality as
\[
S\left(\mathbf{x}^\star + (1-\epsilon)(\mathbf{x}_1 - \mathbf{x}^\star)\right) > S(\mathbf{x}_1).
\]
Since \(\phi(\mathbf{x}_1) = 1\) by assumption, and assuming \(\phi\) remains 1 in a neighborhood where \(S\) does not decrease, we also have
\[
\phi\left(\mathbf{x}^\star + (1-\epsilon)(\mathbf{x}_1 - \mathbf{x}^\star)\right) = 1.
\]
Thus, for \(r = 1 - \epsilon\), we find a point \(0 < r < 1\) such that \(\phi(\mathbf{x}^\star + r(\mathbf{x}_1 - \mathbf{x}^\star)) = 1\), contradicting the assumption that no such \(r\) exists. Therefore, our assumption that \(\cos\alpha_1 < 0\) must be false, and we conclude that
\[
\cos\alpha_1 \geq 0.
\]

Now that we have established \(\cos\alpha_1 \geq 0\), it follows that
\[
\mathbb{E}[\cos \alpha_1] = \mathbb{E}\left[\cos\alpha_1 \,\middle|\, \cos\alpha_1 \geq 0\right].
\]
Expanding \(\cos\alpha_1\) in terms of the vectors involved, we write
\[
\cos\alpha_1 = \left\langle \frac{\mathbf{x}_1 - \mathbf{x}^\star}{\|\mathbf{x}_1 - \mathbf{x}^\star\|_2}, \mathbf{g}_1 \right\rangle.
\]
Thus,
\[
\mathbb{E}[\cos \alpha_1] = \mathbb{E}\left[\left\langle \frac{\mathbf{x}_1 - \mathbf{x}^\star}{\|\mathbf{x}_1 - \mathbf{x}^\star\|_2}, \mathbf{g}_1 \right\rangle \,\middle|\, \left\langle \frac{\mathbf{x}_1 - \mathbf{x}^\star}{\|\mathbf{x}_1 - \mathbf{x}^\star\|_2}, \mathbf{g}_1 \right\rangle \geq 0\right].
\]
Finally, applying the result from \Cref{eqn:init_the_eq} with \(\omega_t = 0\), we obtain
\[
\mathbb{E}\left[\left\langle \frac{\mathbf{x}_1 - \mathbf{x}^\star}{\|\mathbf{x}_1 - \mathbf{x}^\star\|_2}, \mathbf{g}_1 \right\rangle \,\middle|\, \left\langle \frac{\mathbf{x}_1 - \mathbf{x}^\star}{\|\mathbf{x}_1 - \mathbf{x}^\star\|_2}, \mathbf{g}_1 \right\rangle \geq 0\right] = \frac{\Gamma\left(\frac{d}{2}\right)}{2\sqrt{\pi}\,\Gamma\left(\frac{d+1}{2}\right)}.
\]

This completes the proof.

\hfill \(\blacksquare\)
\newpage
\section{Empirical study of AS and AGREST}
\label{app:emp_study}
\begin{figure*}[h!] 
    \centering 
    \small
    \tabcolsep=1.3pt
    \begin{tabular}{@{}c@{}c}
        \includegraphics[width=0.5\textwidth]{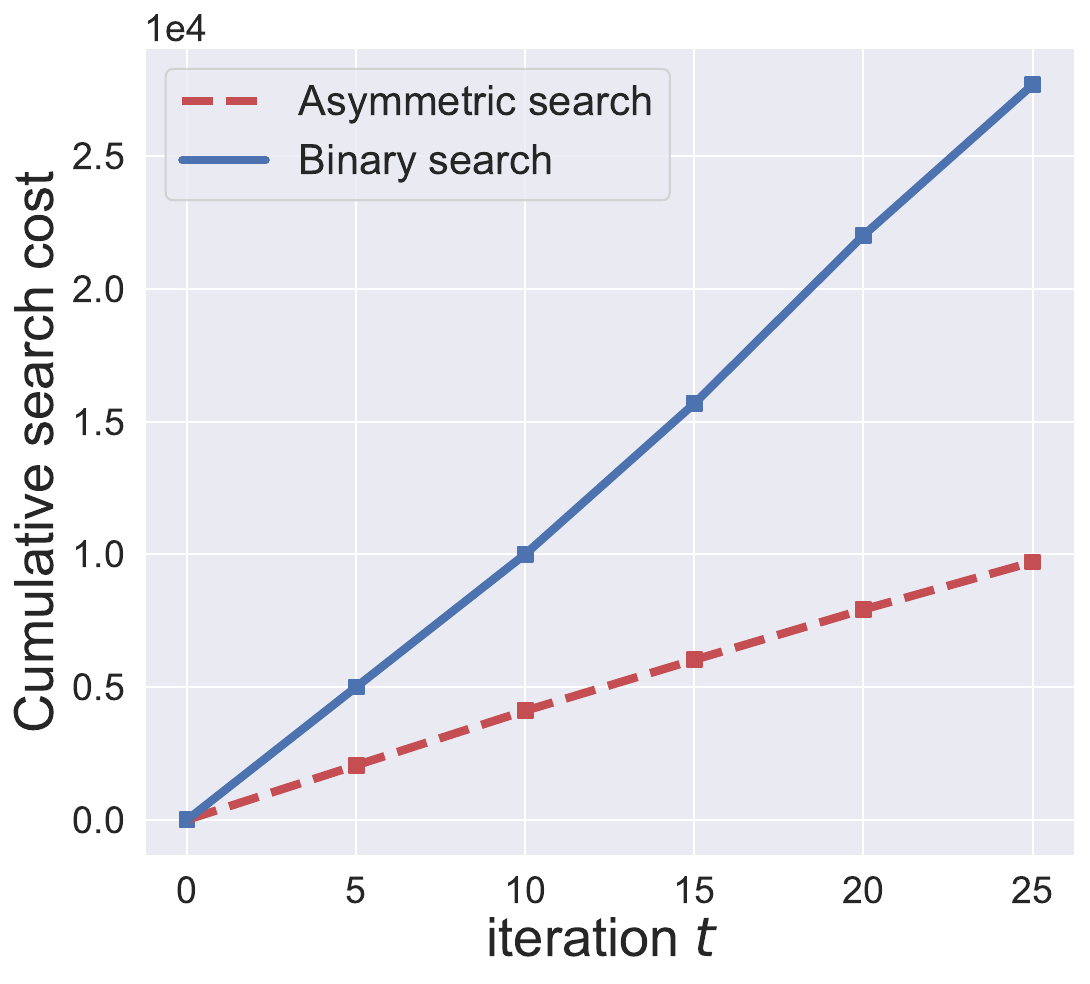} &
        \includegraphics[width=0.5\textwidth]{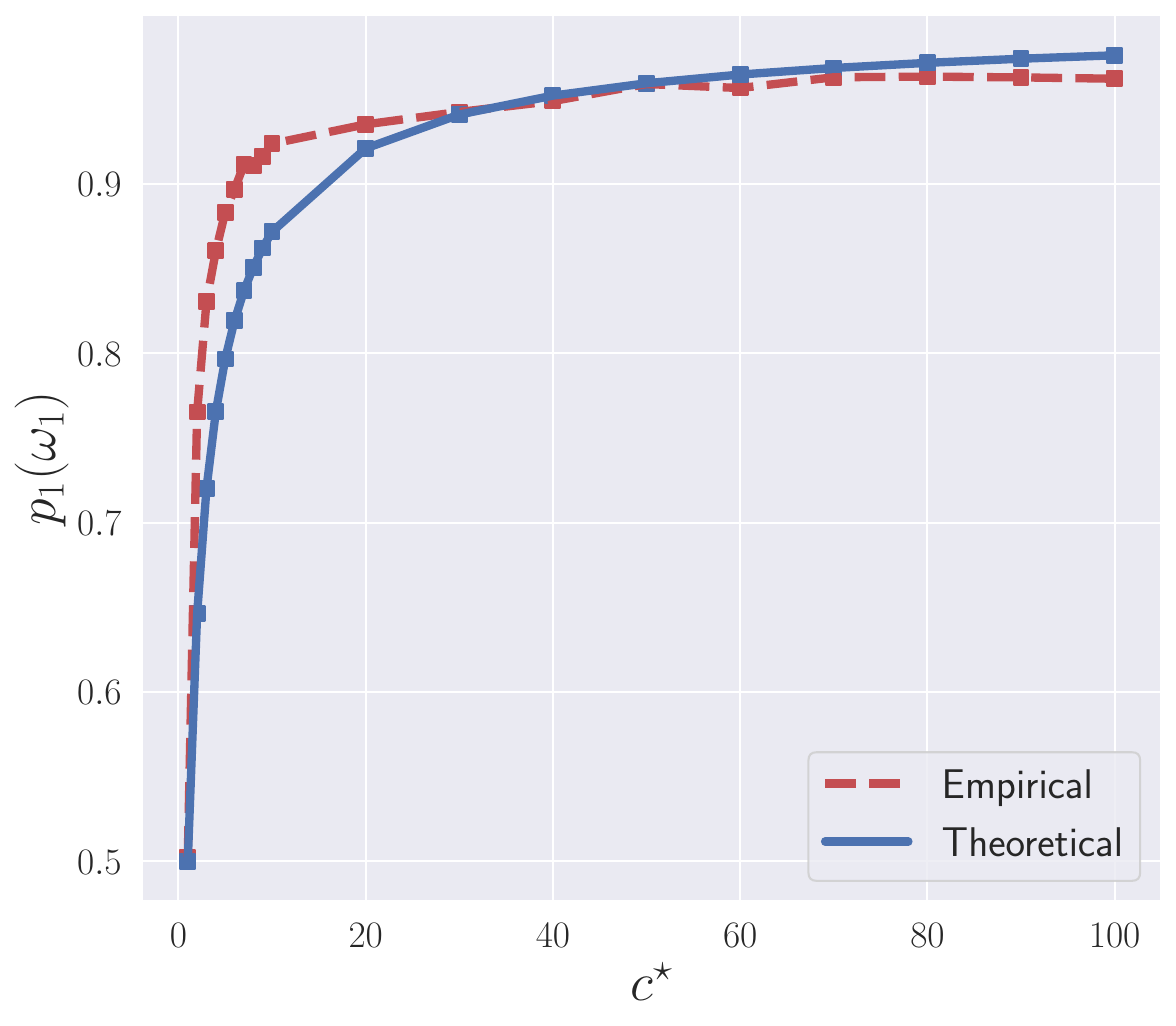}
    \end{tabular}
    \vspace{-2.5mm}
    \caption{Empirical study of AS and AGREST. The left plot compares AS with vanilla search (binary search) in terms of cumulative \emph{search} cost over iterations in GeoDA when \( c^\star = 10^3 \), while the right plot shows the optimal theoretical probability of making low-cost queries (assuming local linearity of the decision boundary) versus the empirical ratio of low-cost to total queries for different values of \( c^\star \).}
    \label{fig:exp_as_agrest}
\end{figure*}

Here, we design two experiments to validate the effectiveness of our analysis for both AS and AGREST. (Fig.~\ref{fig:exp_as_agrest})

First, we sample 100 random correctly classified images from the ImageNet dataset and run the experiments using binary search and asymmetric search when $c^\star = 10^3$. We observe that the average cumulative search cost across iterations for binary search is approximately 2.5 times higher than that of AS. This highlights the effectiveness of AS compared to vanilla search.

Second, to show that using the overshooting value obtained by AGREST leads to a probability of making low-cost queries close to the theoretical value in \Cref{the:sol_obj}, we again sample 100 random images and run one iteration of AGREST using 500 queries for gradient estimation. We then compute the empirical probability of making low-cost queries, defined as the ratio of low-cost to total (500) queries, and compare it to the optimal probability predicted by our theoretical analysis. As shown in Fig.~\ref{fig:exp_as_agrest}, our analysis is close to the empirical results, especially for larger values of $c^\star$.

\newpage
\section{Implementation}
\label{app:imp}
\paragraph{Modification to GeoDA}
\label{app:imp_geoda}

We modify the direction-based adversarial example search phase in GeoDA. In its original implementation, GeoDA estimates the gradient and then proceeds from the original image, taking fixed-size steps along that direction until it finds an adversarial example. However, this process often leads to a large number of flagged queries, since many of the intermediate steps can cross the decision boundary.

To address this issue, we change the starting point of the search. Instead of beginning at the original image, we start from  
\[
\mathbf{x}^{\prime\prime} = \mathbf{x}^\star + \|\mathbf{x}^\star - \mathbf{x}_t\|_2 \cdot \frac{\widehat{\nabla} S(\mathbf{x}_t, \omega_t)}{\|\widehat{\nabla} S(\mathbf{x}_t, \omega_t)\|_2}.
\]
This new starting point lies further in the direction of the estimated gradient and is designed with the expectation that it is already adversarial—or at least closer to an adversarial example than the original image. If \( \mathbf{x}^{\prime\prime} \) is not adversarial, the algorithm continues the search in the estimated gradient direction. This modification significantly reduces the number of flagged queries encountered during the search.

\begin{figure}[h!]
    \centering
    \includegraphics[width=0.5\textwidth]{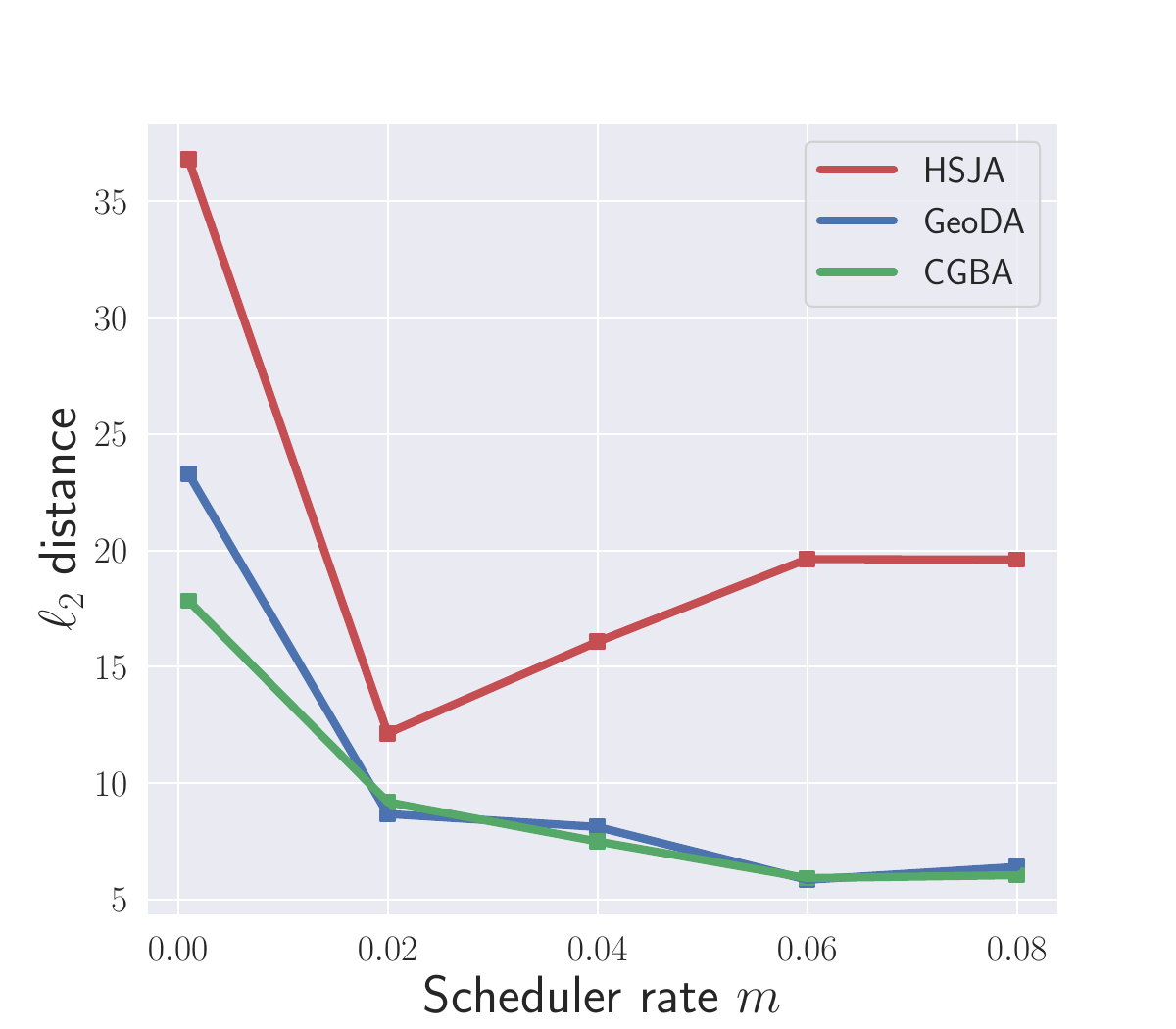}
    \caption{Median \( \ell_2 \) distance of adversarial perturbations for varying values of \( m \), with \( c^\star = 10^3 \) and a total query cost of 150K.}
    \label{fig:m_hyperparam}
\end{figure}

\paragraph{Selection of the hyperparameter \( m \)}
\label{app:imp_hyperparam}
We select values for the hyperparameter \( m \) by evaluating the performance of the corresponding attacks under different settings of \( m \), using 20 randomly selected correctly classified images. This evaluation is performed with \( c^\star = 10^3 \) and a total query cost of 250K, as shown in Fig.~\ref{fig:m_hyperparam}.

Although the optimal value of \( m \) can vary with \( c^\star \), we choose to fix \( m \) independently of \( c^\star \). This decision simplifies the attack process and avoids the additional computational overhead of tuning \( m \) for each value of \( c^\star \), while still enabling effective attack performance.
\paragraph{AGREST with dimension reduction}
\label{app:imp_dim_red}
As mentioned earlier, most practical attacks use a dimension reduction matrix \( \mathbf{R} \in \mathbb{R}^{d \times d'} \) to perform the sampling process in a subspace of dimension \( d' \ll d \), where \( d \) is the dimension of the original space, in order to increase sample efficiency. To apply the same subspace in the AGREST estimator, the only modifications needed compared to the original AGREST~\Cref{alg:age-algorithm} are: first, projecting each sample into the subspace; and second, using the effective dimension \( d' \) to compute \( \alpha_t \). An overview of this version of AGREST is provided in \Cref{alg:ager-algorithm}.
\begin{algorithm}
    \caption{AGREST Estimation}
    \centering
    \label{alg:ager-algorithm}
    \begin{algorithmic}[1]
    \REQUIRE Iteration $t$, source image $\mathbf{x}^\star$, boundary point $\mathbf{x}_t$, dimension $d$, sampling subspace dimension $d^\prime$, sampling subspace matrix $\mathbf{R}$, high-cost query cost $c^\star$, sampling radius $\delta$, sampling batch size $b$, cosine value $\alpha_t$, vanilla gradient estimation query budget $n^\prime_t$, scheduler rate $m$
    \ENSURE Normalized approximated direction $g_t$, next cosine value $\alpha_{t+1}$
    \STATE $n_L \gets 0,\enspace n_H \gets 0, \enspace \mathbf{v}^+ \gets \vec{0},\enspace \mathbf{v}^- \gets \vec{0},\enspace \hat{c} \gets 0,\enspace \omega^\star \gets \textsc{Overshooting}\left(c^\star\right)$ \hfill $\triangleright$ \Cref{the:sol_obj}
    \STATE $\omega_t \gets \omega^\star/\cos\alpha_t, \enspace c_t \gets n^\prime_t(c^\star+1)/2$
    \WHILE {$\hat{c} < c_t$}
    \STATE $B \gets \left\{\mathbf{R}\mathbf{u}_i / \left\|\mathbf{R}\mathbf{u}_i\right\|\text{ where } \mathbf{u}_i\sim\textsc{Uniform}\left(\mathbb{S}^{d-1}\right)\right\}_{i=1}^{b}$ \hfill $\triangleright$ Dimension reduction
    \FOR {\textbf{each} $\mathbf{u}_i \in B$}
    \IF{$\phi\left(\mathbf{x}_t + \omega_t \frac{\mathbf{x}_t-\mathbf{x^\star}}{\left\|\mathbf{x}_t-\mathbf{x^\star}\right\|_2} + \delta \mathbf{u}_i\right) = 1$}
        \STATE $\mathbf{v}^+ \gets \mathbf{v}^+ + \mathbf{u}_i,\enspace n_L \gets n_L + 1, \enspace \hat{c} \gets \hat{c} + 1$
    \ELSE
        \STATE $\mathbf{v}^- \gets \mathbf{v}^- - \mathbf{u}_i, \enspace n_H \gets n_H + 1, \enspace \hat{c} \gets \hat{c} + c^\star$
    \ENDIF 
    \ENDFOR
    \ENDWHILE
    \STATE $\hat{p}_t \gets n_L / \left(n_L + n_H\right)$
    \STATE $g_t \gets \left(1-\hat{p}_t\right)\mathbf{v}^+ + \hat{p}_t\mathbf{v}^-,\enspace\alpha_{t+1}\gets\textsc{Scheduler-Step}\left(t, \hat{p}_t, m\right)$ \hfill $\triangleright$ \Cref{alg:scheduler-algorithm}
    \RETURN $g_t / \left\|g_t\right\|_2,\enspace\alpha_{t+1}$
\end{algorithmic}
\end{algorithm}
\paragraph{Computation resources}
For our experiments on ResNet-50, we use NVIDIA P100 GPUs. All other experiments, including those involving ViT and CLIP models, are conducted on NVIDIA A100 GPUs to accommodate the higher computational and memory demands of these models.
\newpage
\section{Additional results for Vision Transformers (ViT)}
\label{app:ViT_additional_results}
In this section, we present comprehensive empirical evaluations that extend our analysis across varying budget constraints and different query cost parameters $c^\star$. Specifically, we conduct experiments utilizing Vision Transformer architectures (ViT-B/32 and ViT-B/16) on the ImageNet dataset. 

\begin{table}[!htbp]
	\caption{Median $\ell_2$ distance for various $c^\star$ values and different types of attacks for \textbf{ViT-B/32} model on ImageNet dataset.}
	\vspace{2mm}
	\label{t:results_table_ViT_B_32}
	\centering
	\resizebox{\textwidth}{!}{
		\begin{tabular}{|c|c|ccccc|ccccc|ccccc|ccccc|cc|}
			\hline
			\multirow{3}{*}{Attack} & \multirow{3}{*}{Method} & \multicolumn{5}{c|}{$c^\star = 2.0$} & \multicolumn{5}{c|}{$c^\star = 5.0$} & \multicolumn{5}{c|}{$c^\star = 100.0$} & \multicolumn{5}{c|}{$c^\star = 1000.0$} & \multicolumn{2}{c|}{Higher Queries} \\
			\cline{3-24}
			&  & \multicolumn{5}{c|}{Total Cost} & \multicolumn{5}{c|}{Total Cost} & \multicolumn{5}{c|}{Total Cost} & \multicolumn{5}{c|}{Total Cost} & \multicolumn{2}{c|}{Total Cost} \\
			\cline{3-24}
			&  & 1000 & 2000 & 5000 & 10000 & 15000 & 1000 & 2000 & 5000 & 10000 & 15000 & 1000 & 2000 & 5000 & 10000 & 15000 & 1000 & 2000 & 5000 & 10000 & 15000 & 150000 & 250000 \\
			\hline
			\multirow{2}{*}{SURFREE} 
			& VA  & 9.3 & 6.2 & 3.9 & 2.9 & 2.5 & 14.3 & 9.9 & 5.7 & 4.4 & 3.4 & 70.4 & 71.0 & 34.7 & 21.3 & 18.2 & \textbf{69.7} & 71.1 & \textbf{70.6} & 68.1 & \textbf{68.3} & 5.13 & 16.12 \\
			& VA+AS (A-SurFree)  & \textbf{7.8} & \textbf{5.0} & \textbf{3.6} & \textbf{2.4} & \textbf{2.1} & \textbf{11.4} & \textbf{7.3} & \textbf{4.6} & \textbf{3.1} & \textbf{2.5} & \textbf{70.0} & \textbf{70.1} & \textbf{22.8} & \textbf{14.5} & \textbf{9.7} & 70.1 & \textbf{69.9} & 71.5 & \textbf{67.3} & 68.5 & 3.68 & 6.35 \\
			\hline
			\multirow{4}{*}{HSJA} 
			& VA  & 53.7 & 44.8 & 28.7 & 18.3 & 13.9 & 61.6 & 53.2 & 40.0 & 28.1 & 22.8 & \textbf{68.4} & \textbf{67.2} & 67.5 & 67.9 & 63.1 & 69.9 & 71.1 & 69.3 & 69.4 & 70.2 & 4.21 & 22.46 \\
			& VA+AS  & 53.3 & 41.8 & 25.2 & 18.3 & 13.5 & 62.6 & 50.5 & 38.6 & 24.4 & 18.5 & 69.5 & 68.4 & 69.7 & 58.8 & 56.8 & 68.7 & \textbf{69.4} & 70.2 & 68.1 & \textbf{68.2} & 3.88 & 18.79 \\
			& VA+AGREST  & 18.6 & 9.7 & \textbf{4.2} & 2.7 & 2.2 & 36.3 & \textbf{17.3} & 7.8 & 4.4 & \textbf{3.1} & 69.1 & 71.1 & 73.7 & \textbf{37.0} & 35.4 & \textbf{68.5} & 72.1 & \textbf{69.1} & 70.1 & 69.7 & 2.19 & 11.28 \\
			& VA+AS+AGREST (A-HSJA)  & \textbf{17.8} & \textbf{9.6} & 4.7 & \textbf{2.7} & \textbf{2.1} & \textbf{34.8} & 19.1 & \textbf{7.6} & \textbf{4.2} & 3.1 & 71.4 & 69.1 & \textbf{60.1} & 38.3 & \textbf{25.3} & 71.8 & 72.1 & 71.9 & \textbf{67.6} & 68.7 & 2.06 & 10.74 \\
			\hline
			\multirow{4}{*}{GEODA} 
			& VA  & 14.2 & 6.6 & 3.4 & 2.7 & 2.4 & 18.9 & 12.0 & 6.7 & 3.5 & 3.2 & \textbf{67.1} & \textbf{68.2} & \textbf{30.1} & 26.3 & 18.7 & 69.3 & \textbf{68.6} & 68.1 & \textbf{67.5} & 69.2 & 3.97 & 9.83 \\
			& VA+AS  & 12.7 & 7.8 & 3.7 & 2.7 & 2.3 & 17.1 & 11.7 & 5.5 & 3.4 & 2.9 & 71.0 & 73.7 & 30.5 & 20.5 & 16.0 & 70.5 & 71.5 & 68.7 & 73.1 & 71.4 & 3.12 & 8.78 \\
			& VA+AGREST  & \textbf{7.4} & \textbf{3.7} & 2.4 & \textbf{1.9} & \textbf{1.7} & \textbf{14.6} & 8.8 & 4.5 & 3.3 & \textbf{2.7} & 70.2 & 69.8 & 36.6 & 23.5 & 17.7 & \textbf{68.6} & 72.2 & \textbf{66.0} & 69.0 & 69.5 & 2.10 & 5.12 \\
			& VA+AS+AGREST (A-GeoDA)  & 7.8 & 3.9 & \textbf{2.3} & 1.9 & 1.8 & 14.8 & \textbf{8.4} & \textbf{4.4} & \textbf{3.1} & 2.8 & 68.8 & 70.6 & 34.0 & \textbf{19.0} & \textbf{15.6} & 72.2 & 72.5 & 73.5 & 72.9 & \textbf{68.2} & 2.03 & 4.35 \\
			\hline
			\multirow{4}{*}{CGBA} 
			& VA  & 11.6 & 6.0 & 3.0 & 1.6 & 1.4 & 18.4 & 11.3 & 5.3 & 3.1 & 2.1 & 70.4 & 72.0 & 28.0 & 24.5 & 17.0 & \textbf{68.8} & \textbf{70.4} & 69.7 & 72.8 & 72.8 & 2.13 & 9.67 \\
			& VA+AS  & 10.8 & 6.4 & 3.2 & 1.7 & 1.3 & 16.2 & 10.7 & 4.6 & \textbf{2.6} & \textbf{2.0} & 68.5 & 73.4 & 29.2 & 18.1 & 13.7 & 71.9 & 73.8 & 68.3 & \textbf{68.5} & 70.1 & 1.97 & 8.24 \\
			& VA+AGREST  & \textbf{7.3} & \textbf{3.6} & 2.0 & 1.5 & 1.3 & 15.8 & 8.3 & \textbf{4.0} & 3.0 & 2.5 & \textbf{67.1} & 73.5 & 39.2 & 18.9 & 15.2 & 70.9 & 70.5 & 72.7 & 69.8 & \textbf{69.5} & 1.56 & 5.46 \\
			& VA+AS+AGREST (A-CGBA)  & 7.6 & 4.1 & \textbf{1.9} & \textbf{1.5} & \textbf{1.2} & \textbf{14.4} & \textbf{7.7} & 4.2 & 2.9 & 2.5 & 70.2 & \textbf{67.2} & \textbf{23.9} & \textbf{15.9} & \textbf{13.4} & 71.9 & 70.4 & \textbf{67.3} & 70.0 & 72.7 & 1.42 & 5.61 \\
			\hline
	\end{tabular}}
\end{table}
\begin{table}[!htbp]
	\caption{Median $\ell_2$ distance for various $c^\star$ values and different types of attacks  for \textbf{ViT-B/16} model on ImageNet dataset.}
	\vspace{2mm}
	\label{t:results_table_ViT_B_16}
	\centering
	\resizebox{\textwidth}{!}{
		\begin{tabular}{|c|c|ccccc|ccccc|ccccc|ccccc|}
			\hline
			\multirow{3}{*}{Attack} & \multirow{3}{*}{Method} & \multicolumn{5}{c|}{$c^\star = 2.0$} & \multicolumn{5}{c|}{$c^\star = 5.0$} & \multicolumn{5}{c|}{$c^\star = 100.0$} & \multicolumn{5}{c|}{$c^\star = 1000.0$} \\
			\cline{3-22}
			&  & \multicolumn{5}{c|}{Total Cost} & \multicolumn{5}{c|}{Total Cost} & \multicolumn{5}{c|}{Total Cost} & \multicolumn{5}{c|}{Total Cost} \\
			\cline{3-22}
			&  & 1000 & 2000 & 5000 & 10000 & 15000 & 1000 & 2000 & 5000 & 10000 & 15000 & 1000 & 2000 & 5000 & 10000 & 15000 & 1000 & 2000 & 5000 & 10000 & 15000 \\
			\hline
			\multirow{2}{*}{SURFREE} 
			& VA  & 10.7 & 7.1 & 4.2 & 3.0 & 2.3 & 16.6 & 11.0 & 6.8 & 4.4 & 3.6 & \textbf{57.6} & \textbf{57.1} & 33.3 & 20.7 & 18.6 & 60.0 & 56.0 & 56.8 & \textbf{56.8} & \textbf{57.6} \\
			& VA+AS (A-SurFree)  & \textbf{8.9} & \textbf{6.0} & \textbf{3.7} & \textbf{2.3} & \textbf{2.0} & \textbf{14.4} & \textbf{8.3} & \textbf{4.6} & \textbf{3.2} & \textbf{2.6} & 58.2 & 58.2 & \textbf{27.7} & \textbf{15.0} & \textbf{10.4} & \textbf{58.0} & \textbf{54.9} & \textbf{54.6} & 56.9 & 58.3 \\
			\hline
			\multirow{4}{*}{HSJA} 
			& VA  & 37.5 & 29.9 & 17.8 & 10.8 & 7.1 & 49.6 & 40.0 & 26.0 & 18.4 & 13.9 & 57.3 & \textbf{56.0} & 55.1 & 52.5 & 49.7 & 56.7 & 58.9 & 59.2 & 57.2 & 58.0 \\
			& VA+AS  & 38.1 & 28.6 & 18.5 & 9.6 & 6.8 & 47.8 & 38.5 & 23.9 & 16.5 & 11.0 & 57.4 & 56.6 & 55.0 & 47.4 & 44.5 & 56.6 & 60.9 & 57.9 & 59.2 & \textbf{57.4} \\
			& VA+AGREST  & \textbf{14.5} & 8.0 & 4.0 & \textbf{2.3} & \textbf{1.7} & 29.4 & 14.6 & \textbf{6.0} & 3.8 & 2.7 & \textbf{56.7} & 56.9 & 59.6 & \textbf{30.5} & 31.6 & \textbf{55.9} & \textbf{55.3} & 58.2 & 57.4 & 59.0 \\
			& VA+AS+AGREST (A-HSJA)  & 15.1 & \textbf{7.7} & \textbf{3.9} & 2.3 & 1.8 & \textbf{27.1} & \textbf{14.5} & 6.4 & \textbf{3.6} & \textbf{2.6} & 58.1 & 58.1 & \textbf{40.5} & 31.3 & \textbf{21.0} & 58.2 & 56.7 & \textbf{57.5} & \textbf{56.7} & 58.7 \\
			\hline
			\multirow{4}{*}{GEODA} 
			& VA  & 12.5 & 7.3 & 3.1 & 2.2 & 1.8 & 19.7 & 13.7 & 6.5 & 3.3 & 2.5 & 56.8 & 57.7 & 32.5 & 25.5 & 22.6 & 58.2 & \textbf{56.5} & 60.3 & \textbf{57.4} & \textbf{55.8} \\
			& VA+AS  & 14.4 & 7.1 & 3.2 & 2.0 & 1.9 & 19.9 & 12.6 & 5.5 & 3.3 & 2.5 & 60.2 & 60.6 & 30.6 & 23.1 & 17.6 & \textbf{58.0} & 57.0 & \textbf{58.0} & 59.3 & 57.3 \\
			& VA+AGREST  & 8.4 & 4.0 & \textbf{2.1} & \textbf{1.5} & 1.4 & \textbf{13.5} & \textbf{8.1} & \textbf{3.8} & 2.6 & \textbf{2.1} & 58.3 & \textbf{55.8} & 35.5 & 19.4 & 17.0 & 58.8 & 59.0 & 58.4 & 57.6 & 57.4 \\
			& VA+AS+AGREST (A-GeoDA)  & \textbf{7.9} & \textbf{3.8} & 2.1 & 1.6 & \textbf{1.4} & 13.7 & 8.5 & 4.1 & \textbf{2.5} & 2.2 & \textbf{55.9} & 58.5 & \textbf{27.5} & \textbf{18.1} & \textbf{15.2} & 59.2 & 60.6 & 58.4 & 57.7 & 58.0 \\
			\hline
			\multirow{4}{*}{CGBA} 
			& VA  & 11.3 & 6.0 & 2.3 & 1.3 & 1.0 & 16.6 & 11.2 & 4.6 & 2.4 & 1.6 & 57.1 & \textbf{54.9} & 32.4 & 23.7 & 18.6 & \textbf{56.1} & 56.7 & 59.5 & 61.3 & 59.2 \\
			& VA+AS  & 11.2 & 5.6 & 2.4 & 1.3 & 1.0 & 16.4 & 9.5 & 4.3 & 2.2 & \textbf{1.5} & 56.2 & 56.6 & 28.1 & 17.2 & 14.8 & 59.0 & 58.0 & 59.9 & 58.3 & 55.5 \\
			& VA+AGREST  & \textbf{6.5} & \textbf{3.4} & 1.8 & \textbf{1.2} & \textbf{0.9} & 14.0 & 7.9 & 3.8 & 2.2 & 1.8 & \textbf{55.0} & 55.7 & 35.0 & 19.5 & 14.1 & 56.3 & \textbf{55.4} & 59.3 & \textbf{57.6} & \textbf{53.5} \\
			& VA+AS+AGREST (A-CGBA)  & 7.6 & 3.5 & \textbf{1.7} & 1.2 & 1.0 & \textbf{12.2} & \textbf{7.5} & \textbf{3.4} & \textbf{2.2} & 1.9 & 58.1 & 60.5 & \textbf{21.5} & \textbf{16.5} & \textbf{12.2} & 58.2 & 58.2 & \textbf{55.9} & 58.8 & 57.2 \\
			\hline
	\end{tabular}}
\end{table}
\newpage
\section{Asymmetric Search (AS) illustration}
\label{app:AS_illus}
\begin{figure}[!htbp]
    \centering
    \includegraphics[width=1\linewidth, clip]{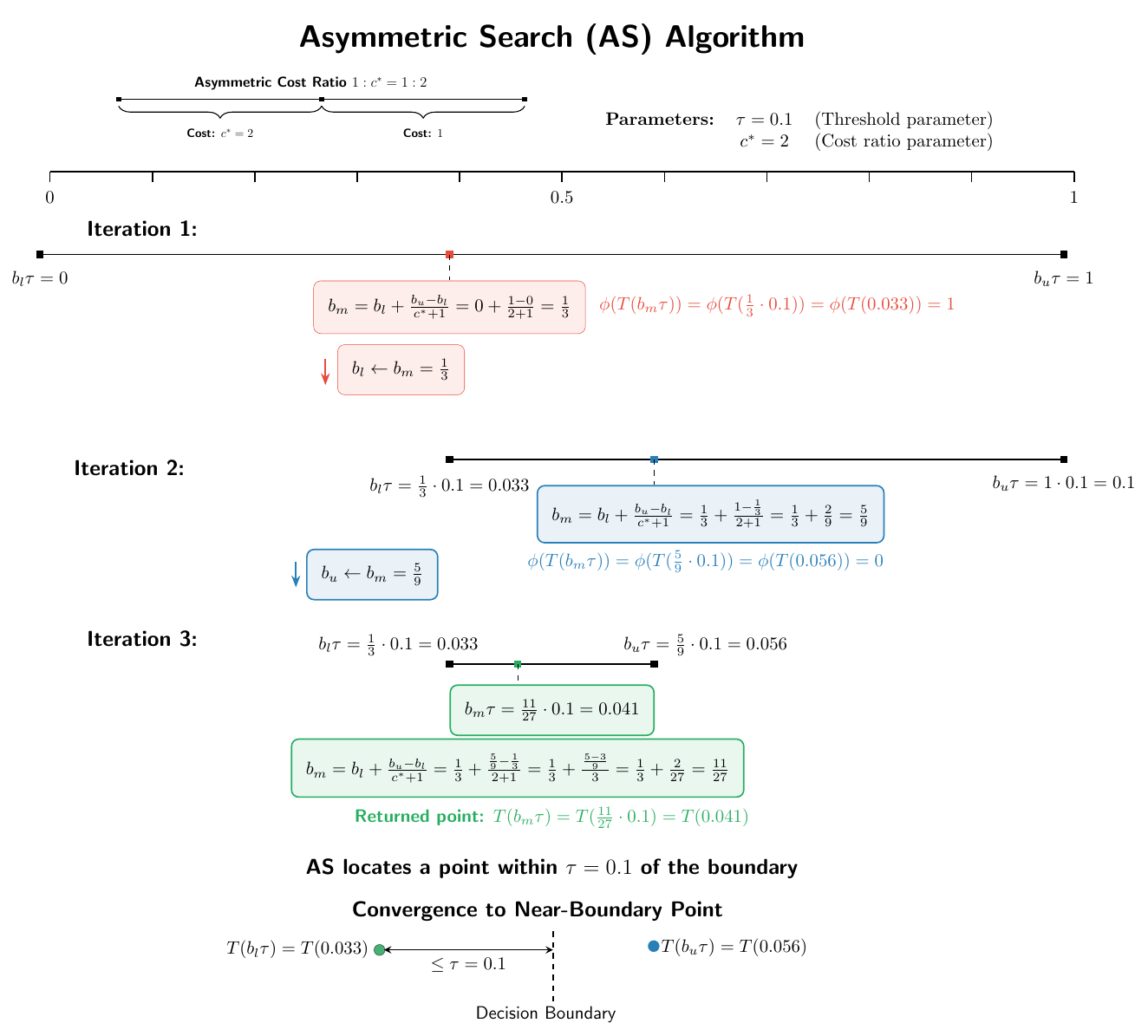}
    \caption{Asymmetric Search (AS) illustration.}
    \label{fig:AS_illus}
\end{figure}

Fig.~\ref{fig:AS_illus} provides a visual example of the Asymmetric Search (AS) algorithm running with parameters $\tau = 0.1$ and $c^\star = 2$. The illustration shows the iterative progression and query evaluations leading to successful convergence near the decision boundary.

\newpage
\section{Asymmetric attacks against CLIP}
\label{app:CLIP_appendix}
\begin{figure}[!htbp]
    \centering 
    \small
    \setlength{\tabcolsep}{12pt} 
    \begin{tabular}{@{}c@{\hspace{1cm}}c@{}} 
        \textbf{Zero-Shot CLIP} & \textbf{Fine-Tuned CLIP} \\
        \includegraphics[width=0.47\textwidth]{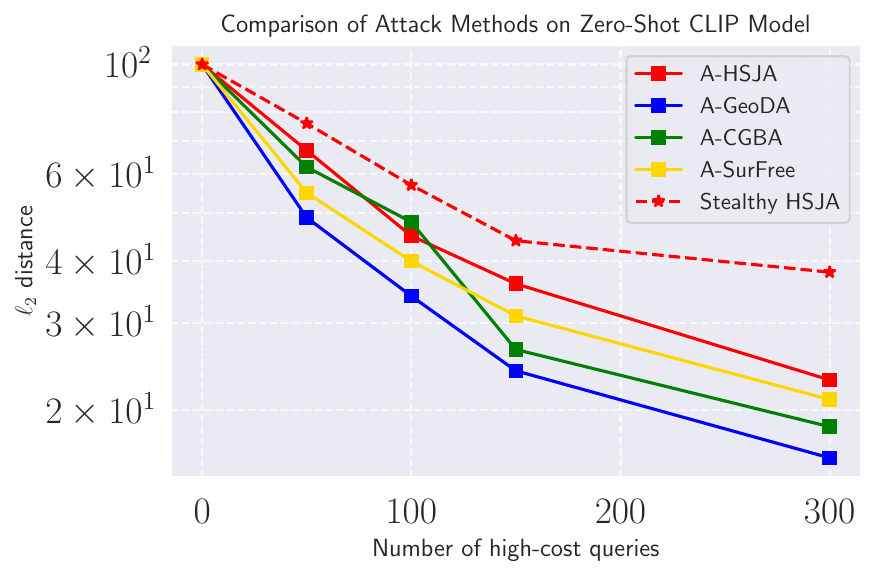} & 
        \includegraphics[width=0.47\textwidth]{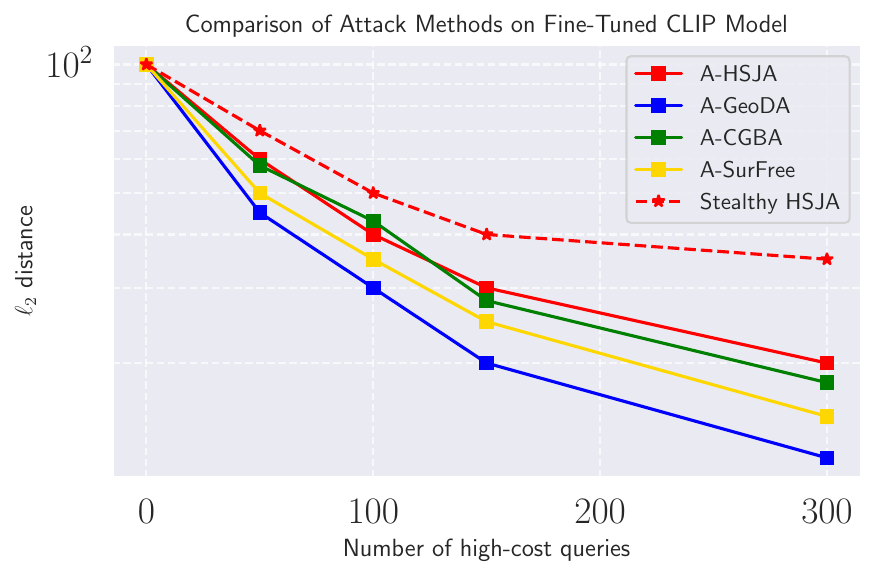}
    \end{tabular}
    \vspace{-2.5mm}
    \caption{Performance of various asymmetric attacks compared to Stealthy HSJA on CLIP.}
    \label{fig:CLIP}
\end{figure}

We evaluate the robustness of vision-language models (VLMs), such as CLIP~\citep{radford2021learningtransferablevisualmodels}, against \textbf{stealthy} adversarial attacks. Our experiments cover both the zero-shot and fine-tuned versions of CLIP. We apply our asymmetric attacks to these models and observe substantial improvements over stealthy baselines. As shown in Fig.~\ref{fig:CLIP}, after making 300 total queries, asymmetric methods achieve 40--60\% lower $\ell_2$ distortion compared to Stealthy HSJA.

\section{Conceptual Illustration}
\label{app:Conceptual_Illustration}
In this section, we show the conceptual illustration of the vanilla gradient estimation and our proposed gradient estimation AGREST.
\begin{figure}[!hbpt]
    \centering
    \includegraphics[width=1\linewidth, clip]{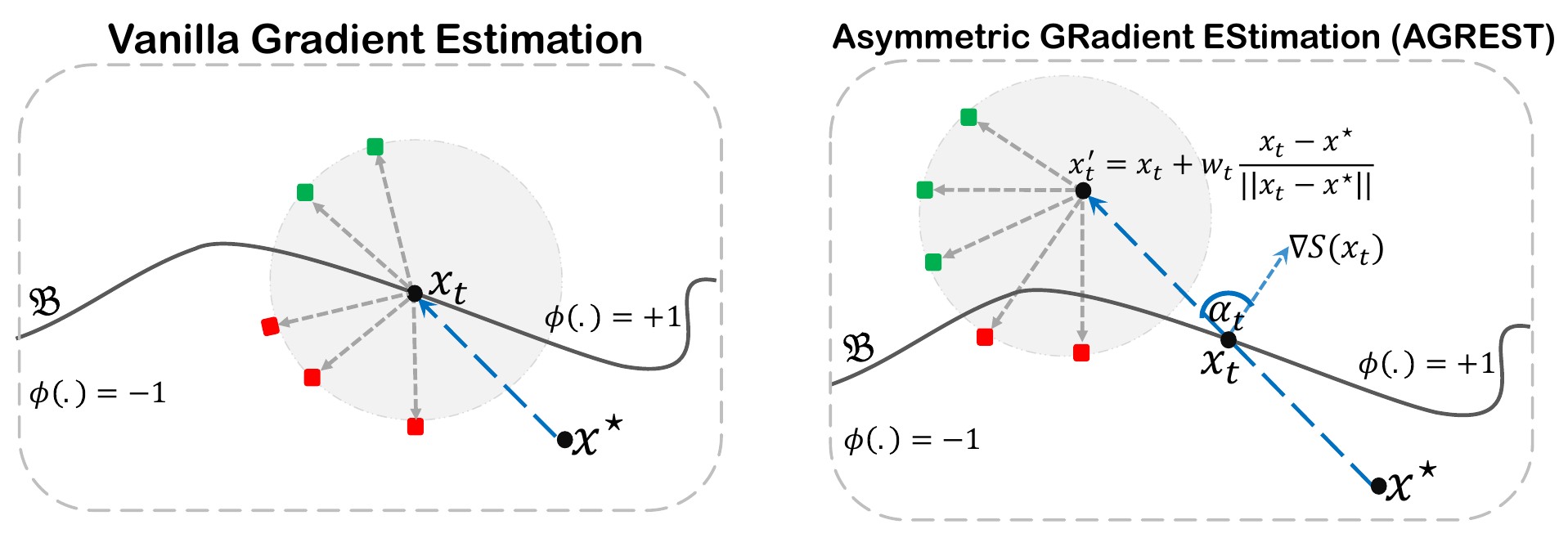}
    \vspace{-0.5cm}
    \caption{\textbf{Comparison of vanilla gradient estimation and its asymmetric counterpart.} Vanilla sampling results in roughly half high-cost and half low-cost queries, whereas AGREST reduces the frequency of high-cost queries by shifting the sampling region and weighting outcomes accordingly.\looseness=-1}
    \label{fig:Conceptual_Illustrations_app}
\end{figure}

\end{document}